%% file: neurips_2026.tex
\documentclass{article}

\usepackage[preprint]{neurips_2026}

\usepackage[utf8]{inputenc} 
\usepackage[T1]{fontenc}    
\usepackage{xcolor}         
\definecolor{mycitecolor}{RGB}{0,0,127} 
\usepackage[colorlinks=true, linkcolor=mycitecolor, urlcolor=mycitecolor, citecolor=mycitecolor]{hyperref}       
\usepackage{url}            
\usepackage{booktabs}       
\usepackage{amsfonts}       
\usepackage{nicefrac}       
\usepackage{microtype}      

\usepackage{float}
\usepackage{pifont}
\usepackage{wrapfig}
\usepackage{amsmath,amssymb,amsfonts,bm}
\usepackage{algorithmic}
\usepackage{graphicx}
\usepackage{textcomp}
\usepackage{siunitx}
\usepackage{soul}
\usepackage{xspace}
\usepackage{booktabs}
\usepackage{multirow}
\usepackage{caption}
\usepackage{balance}
\usepackage[inline]{enumitem}
\usepackage[export]{adjustbox}
\usepackage[ruled,vlined]{algorithm2e}
\usepackage{marvosym} 
\usepackage{subcaption}
\newcommand{\figref}[1]{\hyperref[#1]{Figure~\ref*{#1}}}
\newcommand{\tabref}[1]{\hyperref[#1]{Table~\ref*{#1}}}
\newcommand{\secref}[1]{\hyperref[#1]{\S~\ref*{#1}}}
\newcommand{\appendixref}[1]{\hyperref[#1]{Appendix~\ref*{#1}}}
\newcommand{\equref}[1]{\hyperref[#1]{Equation~\ref*{#1}}}
\newcommand{\algref}[1]{\hyperref[#1]{Algorithm~\ref*{#1}}}

\newcommand{\squishlist}{
\begin{list}{$\bullet$}{
  \setlength{\itemsep}{0pt}
  \setlength{\parsep}{3pt}
  \setlength{\topsep}{3pt}
  \setlength{\partopsep}{0pt}
  \setlength{\leftmargin}{3.5mm}
  \setlength{\labelwidth}{1em}
  \setlength{\labelsep}{0.5em}}}
\newcommand{\squishend}{\end{list}}
\newcommand{\methodname}{{HCInfer}\xspace}

\title{\methodname: An Efficient Inference System via Error Compensation for Resource-Constrained Devices}

\author{%
  Shen Xu$^{1}$, 
  Xiangwen Zhuge$^{1}$, 
  Zhe Xu$^{2}$,
  Yingkun Hu$^{1}$,
  Zheng Yang$^{1}$,
  Yunhao Liu$^{1}$
  \\
  $^1$Tsinghua University \\
  $^2$Huazhong University of Science and Technology \\
}

\begin{document}

\maketitle

\input{body/0-abstract}

\input{body/1-introduction}

\input{body/2-related-work}

\input{body/3-observations}

\input{body/4-method}

\input{body/5-evaluation}

\input{body/6-conclusion}

\newpage
\bibliographystyle{unsrt}  
\bibliography{reference}   

\newpage
\input{body/appedix}

\newpage
\input{body/checklist.tex}

\end{document}

%% file: body/0-abstract.tex
\begin{abstract}
LLMs often struggle with memory-constrained deployment on consumer-grade hardware due to their massive parameter sizes.
While existing solutions such as model compression and offloading improve deployment feasibility, they often suffer from substantial accuracy degradation or severe throughput bottlenecks.
Recent error compensation methods recover accuracy through auxiliary LoRA-style branches, and we observe that these branches are inherently amenable to offloading: they require substantial parameter storage but access only a small subset of compensation parameters during each inference step.
Motivated by this opportunity, we propose \methodname, a heterogeneous inference system that offloads residual compensation to the CPU while executing the compressed backbone on the GPU, and further introduces an asynchronous compensation pipeline and sensitivity-aware dynamic rank allocation to hide compensation overhead and maximize accuracy recovery.
Experimental results show that \methodname achieves a maximum accuracy improvement of 5.2\% on downstream tasks compared to compression model and sustaining a maximum speedup of 10.4x compared to full-precision model.
\end{abstract}

%% file: body/1-introduction.tex
\section{Introduction}
Large language models(LLMs) continue to demonstrate breakthroughs in scenarios such as autonomous scientific \cite{zheng2025automation}, multi-agent collaboration \cite{li2024survey}, and complex decision-making \cite{huang2025foundation}.
However, deploying LLMs with tens of billions of parameters on consumer-grade hardware remains challenging due to limited GPU memory. For example, Qwen3-30B-A3B~\cite{qwen3} requires over 60~GB of memory for parameter storage, far exceeding the 24~GB VRAM capacity of a representative consumer GPU such as the NVIDIA GeForce RTX 4090.

\begin{figure}
    \centering
    \includegraphics[width=0.9\textwidth]{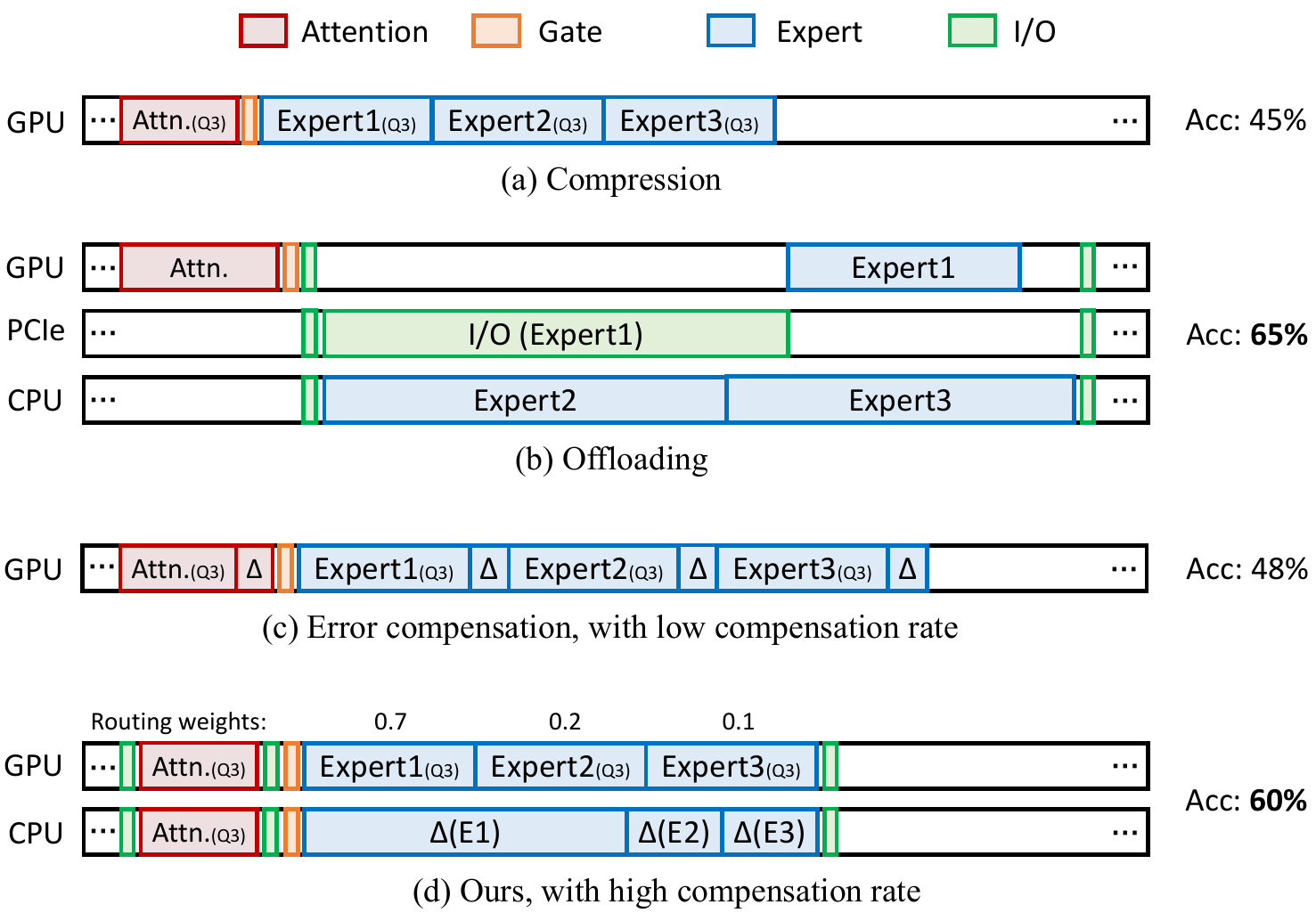}
    \caption{Overview of different inference paradigms.}
    \label{fig:pipeline-overall}
\end{figure}

To improve the accessibility of LLMs on consumer-grade hardware, prior work has followed two main technical directions. (1) Compression methods
trade weight precision for storage reduction and computational acceleration \cite{gptq, awq, efficientqat}, as illustrated in \figref{fig:pipeline-overall}(a). Although such methods can fully utilize GPUs and achieve high inference throughput, model accuracy often drops substantially under extremely aggressive compression ratios. (2) Offloading methods
store full-precision parameters that cannot fit into GPU memory in CPU memory and load them to GPU memory on demand during inference \cite{flexgen, hetegen, powerinfer}. Although offloading preserves the original model accuracy, its throughput is limited by the bandwidth bottleneck of the PCIe bus. Frequent data movement often causes orders-of-magnitude degradation in inference throughput, as shown in \figref{fig:pipeline-overall}(b).

The limitations of existing methods naturally raise the following question: \emph{Can we design a new LLM deployment approach that preserves model accuracy as much as possible while still achieving high inference throughput?}

Recent progress in low-rank quantization residual compensation provides a promising direction to resolve this tension \cite{aser, eora}. Methods observe that quantization error matrices exhibit clear low-rank structure \cite{aser}.
By approximating the quantization error with LoRA-style low-rank matrices,
these methods can repair quantization errors with limited additional computation, thereby improving inference accuracy,
as illustrated in \figref{fig:pipeline-overall}(c). However, when deploying such compensation methods on devices where GPU memory is already highly constrained, the remaining GPU memory can store only a very limited number of compensation parameters, resulting in restricted compensation effectiveness and narrow applicability.

To address this challenge, we observe that the two components of low-rank compensation methods are naturally suitable for offloading techniques. We find that residual compensation is highly memory-intensive: although it depends on large compensation matrices, only a small subset of parameters is accessed during actual computation. This behavior fundamentally differs from the compute-intensive nature of quantized weight computation, and is naturally suitable for CPU-GPU heterogeneous parallel inference. We further observe that different weight matrices exhibit different sensitivities to quantization error compensation.
This suggests that dynamic rank allocation can achieve more effective accuracy recovery than static strategies under the same computation budget. 

Based on these observations, we design and implement \methodname, a heterogeneous inference system with dynamic error compensation, as shown in \figref{fig:pipeline-overall}(d). \methodname\ builds a heterogeneous compensation pipeline that places quantized operators and residual compensation on the GPU and CPU, respectively, and achieves near-zero-overhead compensation through asynchronous parallel execution. In addition, \methodname\ introduces a sensitivity-aware dynamic rank allocation algorithm that adaptively assigns compensation ranks to different matrices. This design enables the system to run quantized models with near full-precision accuracy on consumer-grade hardware while incurring almost no throughput loss. 

In summary, this paper makes the following contributions:

\begin{itemize}[leftmargin=2em, topsep=-2pt, parsep=0pt]
    \item We propose and implement \methodname, a novel heterogeneous inference system based on dynamic error compensation, making high-accuracy and high-throughput deployment of LLMs on consumer-grade hardware feasible.
    
    \item We develop a heterogeneous compensation pipeline.
    \methodname fully exploits the memory-bound nature of residual compensation and achieves substantial temporal overlap between heterogeneous computation.
    
    \item We design a sensitivity-aware dynamic rank allocation algorithm.
    \methodname allocates compensation resources according to multi-dimensional sensitivity indicators to achieve maximum compensation efficiency.
    
    \item Experiments on Qwen-30B-A3B \cite{qwen3} and Llama-3.1-8B \cite{llama} show that, \methodname improves accuracy by 2.3\% to 5.2\% while preserving at least 85\% of throughput compared with quantized baselines, and achieves up to 10.4$\times$ speedup compared with full-precision baselines.
\end{itemize}

%% file: body/2-related-work.tex
\section{Related Work}

\subsection{Quantization}

Model quantization compresses LLMs by converting high-precision weights and activations into low-bit representations.
Existing approaches mainly include post-training quantization (PTQ) and quantization-aware training (QAT). PTQ avoids retraining by calibrating on small datasets \cite{gptq, abq-llm, llmint8, mobilequant}, with methods improving accuracy via channel protection \cite{awq} and residual compensation \cite{decdec}. QAT incorporates quantization noise during training to achieve higher accuracy \cite{efficientqat, dlqat, llmqat}, with recent works improving efficiency through low-rank training \cite{dlqat}, distillation \cite{llmqat}, and block-wise optimization \cite{efficientqat}.
Despite these advances, aggressive low-bit quantization still leads to significant accuracy degradation.

\subsection{Offloading}

Parameter offloading enables large-model inference on limited hardware by dynamically transferring weights or states between GPU and external memory. Early systems \cite{accelerate, deepspeed} focus on pure offloading, while later works introduce hybrid CPU-GPU execution, overlapping computation and I/O to improve efficiency \cite{flexgen, hetegen, powerinfer}. For MoE models, prior methods further exploit expert sparsity via selective loading \cite{fiddler} and mixed-precision compression \cite{hobbit}.
However, offloading performance is fundamentally limited by I/O bandwidth,
restricting throughput in practical deployments.

\subsection{LoRA-Style Quantization Error Compensation}

Recent studies show that quantization errors exhibit low-rank structure, making LoRA-style adapters effective for error compensation \cite{aser, eora}. Prior works leverage low-rank approximation for fine-tuning or direct reconstruction, improving the accuracy of quantized models with varying training and overhead costs \cite{lqlora, loftq, lorc}.

Inspired by these studies, we propose \methodname, a deployment-oriented inference system that adopts LoRA-style compensation for quantization errors while further addressing the practical challenge of executing the compensation branch efficiently under memory-constrained heterogeneous hardware.

%% file: body/3-observations.tex
\section{Observations}


\subsection{Non-Uniform Impact of Quantization Error Across Model Components}
\label{sec:obs-quant}

\begin{wrapfigure}{r}{0.38\textwidth}
    \centering
    \includegraphics[width=0.38\textwidth]{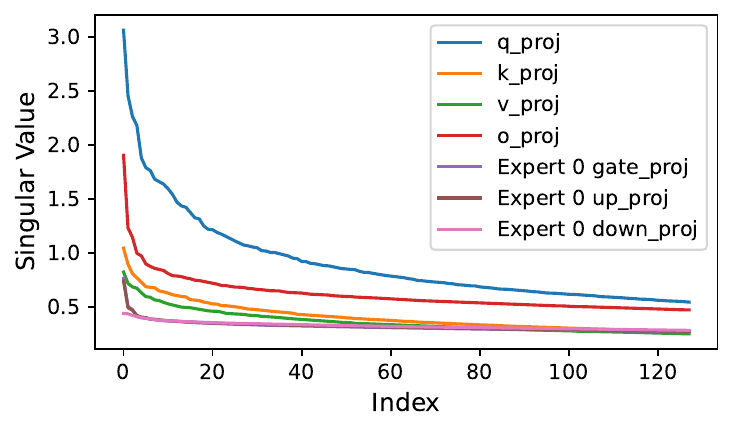}
    \caption{Singular value distribution of quantization error matrices for various modules in layer 0 of Qwen-30B-A3B, based on GPTQ Q4 quantization.}
    \label{fig:obs-sv}
\end{wrapfigure}

Existing studies have shown that quantization error matrices often exhibit strong low-rank structure \cite{aser, eora}. However, our analysis reveals that the usefulness of compensation varies substantially across different components of the model.

\emph{Singular-value distribution.} We examine the singular value spectrum of quantization residual matrices from different operators in Qwen-30B-A3B, shown in \figref{fig:obs-sv}. While many matrices exhibit concentrated spectra, the degree of concentration differs significantly across operators. 
Matrices with more concentrated singular values should receive higher compensation priority for their higher intrinsic recoverability.

\begin{figure}[t]
    \centering
    \includegraphics[width=1.0\textwidth]{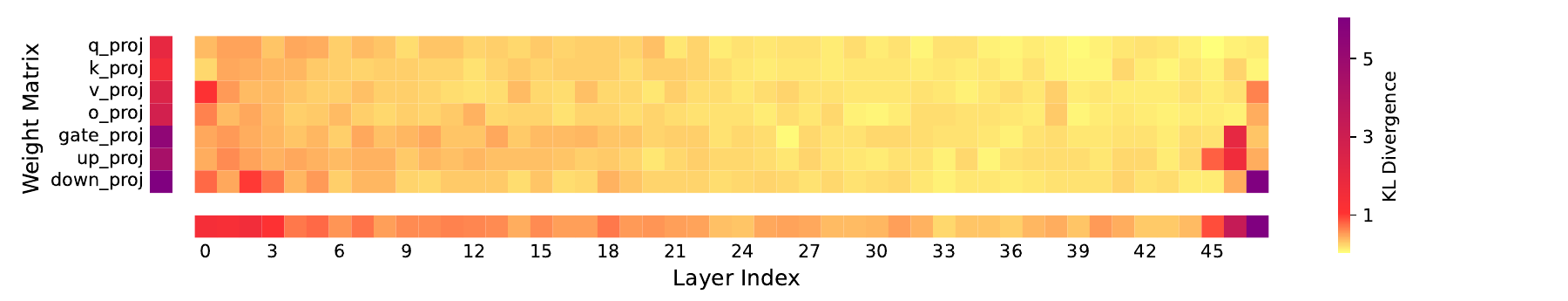}
    \caption{Matrix-wise output sensitivity analysis of Qwen-30B-A3B under GPTQ Q4 quantization. A higher KL divergence value indicates a greater discrepancy between the quantized model’s output distribution and the original full-precision output.}
    \label{fig:obs-kl}
\end{figure}

\emph{Output sensitivity.} By sequentially replacing each weight matrix with its quantized counterpart and measuring the KL divergence between the outputs before and after the replacement, we observe in \figref{fig:obs-kl} that quantization errors introduced at different decoder layers and projections affect final predictions unevenly.
This suggests that compensation strategies should account not only for local matrix recoverability but also for layer-level structural sensitivity.

\emph{Expert weights in MoE models.} For MoE architectures, since the output of an MoE block is a weighted aggregation of activated experts, the benefit of compensating a particular expert is naturally modulated by its routing weight. Experts with larger routing probabilities should receive more compensation resources.

These observations motivate the sensitivity-aware dynamic rank allocation strategy introduced in \secref{sec:method-rank}, where compensation resources are allocated adaptively to maximize accuracy recovery.

\subsection{Compute-Memory Characteristics of Quantized Weights and LoRA Compensation}
\label{sec:hete}

We analyze the distinct computational properties of the quantized backbone and the low-rank compensation branch.

\emph{Quantized backbone is compute-bound.} The primary advantage of low-bit quantization is that it significantly reduces model storage cost \cite{gptq}, thereby increasing arithmetic intensity during inference. As a result, GPUs are well suited for executing the quantized backbone path.

\emph{Low-rank compensation is memory-bound.} Since the full pool of compensation parameters for all matrices in a model can approach the size of the original model \cite{eora}, it is impractical to store them entirely in GPU VRAM.
Despite the large model footprint, only a small subset of these parameters is activated during each inference step, rendering the compensation branch highly memory-bound with low arithmetic intensity.
Given the significantly larger memory capacity of the CPU and the relatively low computational demand of the compensation task, offloading these operations to the CPU is both feasible and highly efficient.

Based on this observation, \methodname decomposes model inference into GPU and CPU pathes, and design efficient heterogeneous inference pipeline in \secref{sec:method-pipeline}.

%% file: body/4-method.tex
\section{\methodname: Heterogeneous LoRA-Style Error Compensation Inference}

\begin{figure}
    \centering
    \includegraphics[width=1.0\textwidth]{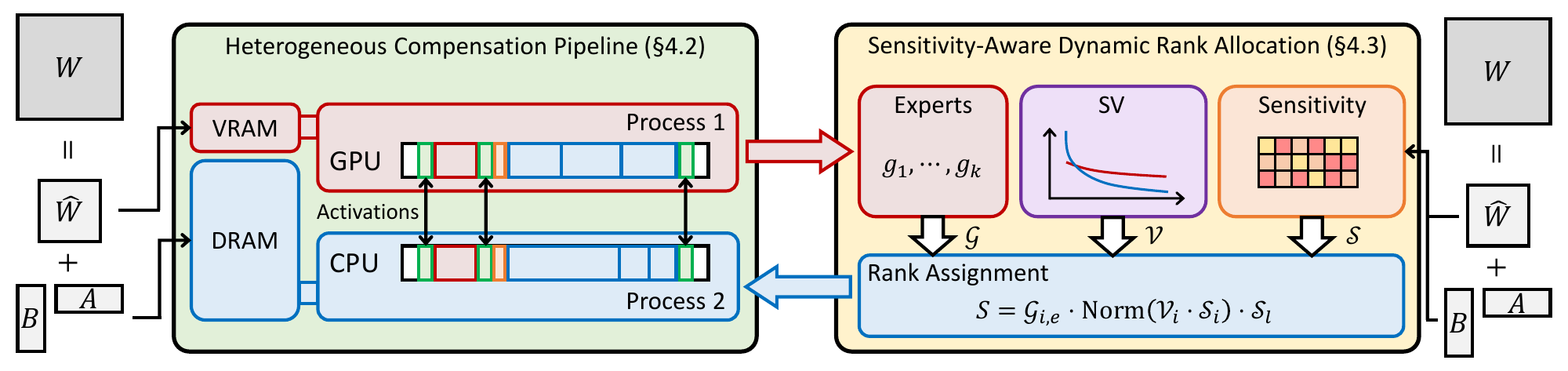}
    \caption{Overview of \methodname.}
    \label{fig:framework}
\end{figure}

In this section, we present \methodname, a deployment framework that improves the inference accuracy of quantized large language models.
\methodname consists of two key components: a heterogeneous compensation pipeline, where the GPU executes the quantized backbone model while the CPU asynchronously computes the residual compensation branch; a sensitivity-aware dynamic rank allocation strategy, which measures the sensitivity of different model components prior to deployment and allocates compensation ranks accordingly. \figref{fig:framework} illustrates the overall design of \methodname.

\subsection{Theoretical Framework}
\label{sec:framework}


Consider a linear layer in an LLM with weight matrix $W\in\mathbb{R}^{d\times k}$ and input tensor $X\in\mathbb{R}^{n\times d}$. Its original full-precision computation is given by $Y=XW$. After low-bit quantization, the weight matrix becomes $\hat{W}$, which introduces a quantization error $\Delta W=W-\hat{W}$.

Based on the observations in \secref{sec:obs-quant}, the quantization error matrix $\Delta W$ exhibits strong low-rank structure, shown as:

\[
Y = X\hat{W} + X\Delta W \approx X\hat{W} + (XA_r)B_r,
\]

where $A_r \in \mathbb{R}^{d \times r}$ and $B_r \in \mathbb{R}^{r \times rk}$ are low-rank factors obtained from the singular value decomposition of $\Delta W$, and $r\ll \min(d,k)$ denotes the compensation rank.


\subsection{Heterogeneous Compensation Pipeline}
\label{sec:method-pipeline}

To make full use of computing resources on resource-constrained devices, \methodname constructs an asynchronous inference pipeline
through \emph{workload decoupling} and \emph{execution flow reorganization}. Detailed implementation of the pipeline is discribed in \appendixref{sec:appendix-pipeline}.

\subsubsection{Workload Decoupling}

As discussed in \secref{sec:hete}, the quantized backbone computation $\hat{Y} = X\hat{W}$ is a compute-bound task suited for the GPU, while the residual compensation $\Delta Y = X\Delta W = (XA)B$ branch's memory-bound nature makes it ideal for CPU offloading.
Accordingly, \methodname decomposes each linear layer into two parallel execution paths:

\begin{itemize}[leftmargin=2em, topsep=-2pt, parsep=0pt]
    \item \textbf{Backbone path, GPU-driven.} The quantized model resides entirely in GPU memory. During execution, the GPU directly invokes optimized low-bit matrix multiplication kernels to compute $\hat{Y} = X\hat{W}$.
    \item \textbf{Compensation path, CPU-driven.} The large LoRA parameter pool is stored in host DRAM. During execution, the GPU transfers intermediate activations to the CPU process and asynchronously triggers compensation computation. The CPU then applies the dynamic ranks and computes $\Delta Y = (XA_r)B_r$, after which the compensation result is sent back to the GPU.
\end{itemize}


To eliminate repeated process launch overhead, \methodname adopts a persistent dual-process architecture. At initialization, two long-lived processes are created in parallel: a GPU process for quantized inference and a CPU process for compensation computation. These processes remain active throughout execution and communicate via shared memory and lightweight signaling primitives.


\subsubsection{Execution Flow Reorganization}

To enable shared compensation resources across multiple operators, \methodname reorganizes the execution flow by merging communication stages and restructuring computation for MoE architectures.

\emph{Communication Fusion.}
\methodname merges communication and calculation according to operator dependencies to enable cross-operator CPU sharing. The attention projection operators $Q(X), K(X), V(X)$ are mutually independent. Therefore, \methodname logically groups these projections at the beginning of the layer. The GPU sends the input activation $X$ only once, after which the CPU computes three compensation branches in parallel $\Delta Y_{Q,K,V} = (XA_{Q,K,V})B_{Q,K,V}$. The three compensation outputs are then returned to the GPU in a single synchronized transfer. For the feed-forward block, \methodname applies the same strategy and fuses communication for the compensation branches associated with $W_{\text{up}}$ and $W_{\text{gate}}$.




\emph{Cross-Expert Pipeline Reorganization.}
\methodname restructures standard MoE inference workflow by grouping identical operators across all activated experts.
After router execution on the GPU, the GPU and CPU jointly process all activated experts' $W_{\text{up}}$ and $W_{\text{gate}}$ operators in parallel. Once compensation results are synchronized, nonlinear activation is applied for all experts on the GPU. The system then jointly processes all $W_{\text{down}}$ operators before aggregating expert outputs.
The GPU transmits intermediate activations once at the start of the $W_{\text{up}}, W_{\text{gate}}$ stage and once at the start of the $W_{\text{down}}$ stage. The CPU returns compensation outputs only once after each stage.




By grouping homogeneous operators, \methodname allows multiple matrices to dynamically share CPU compensation capacity within the same execution window while reducing communication frequency.

\subsection{Sensitivity-Aware Dynamic Rank Allocation}
\label{sec:method-rank}

To maximize accuracy recovery under limited CPU compute throughput and memory bandwidth,
\methodname dynamically allocates rank budgets according to (i) the intrinsic recoverability of the quantization error, (ii) matrix- and layer-level output sensitivity, and (iii) expert activation for MoE models. Details of the algorithm are discribed in \appendixref{sec:appendix-algorithm}.

\subsubsection{Intrinsic Recoverability}

For a quantization residual matrix $\Delta W_i \in \mathbb{R}^{d_i \times k_i}$,
its singular value decomposition is $\Delta W_i = U_i \Sigma_i V_i^\top$, where $\Sigma_i = \operatorname{diag}(\sigma_{i,1}, \sigma_{i,2}, \dots, \sigma_{i,n_i}),\ \sigma_{i,1} \geq \sigma_{i,2} \geq \cdots \geq \sigma_{i,n_i} \geq 0$.
We partition the singular values into a salient set $S_i$ and a residual set $R_i$ by maximizing difference in their variance.

We then define the singular value salience score of matrix $i$ as
\[
\phi_i =
\frac{\frac{1}{|S_i|}\sum_{\sigma\in S_i}\sigma}
{\frac{1}{|R_i|}\sum_{\sigma\in R_i}\sigma},
\]

Within each layer $\ell$, these scores are normalized by layer to obtain relative recoverability $\mathcal{V}_i = \mathrm{Norm}(\phi_i)$.

\subsubsection{Multi-Dimensional Output Sensitivity}

Although $V_i$ captures how efficiently a matrix can be approximated, it does not reflect the impact of that matrix on the final model output. We first measure the importance of each matrix by individually replacing it with its quantized counterpart and measure KL divergence between output distributions:
\[
D_i = D_{\mathrm{KL}}\bigl(P(Y|X)\,\|\,Q_i(Y|X)\bigr).
\]
We then therefore introduce two complementary sensitivity measurements.

\emph{Matrix-Level Sensitivity.}
For each layer $\ell$, we normalize the KL divergence among matrices within that layer as matrix-level sensitivity score $\mathcal{S}_i = \mathrm{Norm}(D_i)$.



\emph{Layer-Level Sensitivity.} Different transformer layers contribute unequally to final model quality. To capture this effect, we take a similar approach to matrix-level sensitivity measurement by replacing all matrices in layer $\ell$ with their quantized versions, and measure the KL divergence $D_{\ell}$.




To emphasize highly sensitive layers within the limited compute budget, we identify the set of top-K most sensitive layers, denoted as $\mathcal{T}$, and derive the layer-level sensitivity score $S_\ell$ as follows:

\[
\mathcal{S}_\ell =
\begin{cases}
1, & \ell \in \mathcal{T}, \\
\frac{D_\ell}{\min_{m \in \mathcal{T}} D_m}, & \ell \notin \mathcal{T}.
\end{cases}
\]


\subsubsection{Expert Activation for MoE Models}

For MoE models, \methodname further reallocates compensation ranks using routing weights. Suppose the top-$k$ activated experts are indexed by $e\in\mathcal{A}$, with routing probabilities $g_e$.
Computational resource is shared among the activated $k$ experts, therefore the expert activation score $\mathcal{G}_{i,e}$ is scaled by $\mathcal{G}_{i,e}=k\cdot g_e$.



\subsubsection{Priority Score and Rank Allocation}

We combine all three factors into a unified compensation priority score $S$. For matrix $i$ in layer $\ell$, the score is defined as

\[
\mathcal{P}_i = \mathrm{Norm}(\mathcal{V}_i\cdot \mathcal{S}_i)\cdot \mathcal{S}_\ell.
\]


For MoE models, the score is further scaled by activation score $\mathcal{G}_{i,e}$:

\[
\mathcal{P}_{i,e} = \mathcal{G}_{i,e}\cdot \mathrm{Norm}(\mathcal{V}_i\cdot \mathcal{S}_i)\cdot \mathcal{S}_\ell.
\]


Next, we determine a hardware-dependent standard rank $r_{\mathrm{std}}$, defined as the maximum rank whose CPU computation and communication latency can still be hidden behind GPU execution:

\[
r_{\mathrm{std}}
=
\max_r
\left\{
T_{\mathrm{CPU}}(r)+T_{\mathrm{comm}}
\le
T_{\mathrm{GPU}}
\right\}.
\]

Finally, the allocated compensation rank for matrix $i$ is given by $r_i = \left\lfloor \mathcal{P}_i \cdot r_{\mathrm{std}} \right\rfloor$. The proof of the optimality of the algorithm is given in \appendixref{sec:appendix-proof}.



%% file: body/5-evaluation.tex
\section{Evaluation}

\subsection{Experimental Setup}
\label{sec:setup}

\textbf{Implementation:} We implemented \methodname based on the Hugging Face Transformers (v4.57.1) \cite{huggingface}. GPTQ \cite{gptq} was selected as our fundamental quantization method.

\textbf{Hardware Environment:} Experiments were conducted on NVIDIA RTX 4090 with 24GB VRAM and Intel i9-10980XE with 256GB DRAM, interconnected via PCIe Gen3 x16.

\textbf{Models and Quantization Configuration:} We selected Qwen-30B-A3B \cite{qwen3}, a large-scale MoE model, and Llama-3.1-8B \cite{llama}, a small-to-medium scale dense model, for implementation. Our method was implemented based on a 4-bit and 3-bit quantized version respectively. For Llama-3.1-8B, we limited the available GPU VRAM to 8GB to simulate an extremely constrained environment.

\textbf{Baselines:} We compared \methodname against two categories of mainstream inference deployment solutions:
\begin{enumerate}[leftmargin=2em, topsep=-2pt, parsep=0pt]
    \item \textbf{Pure Quantized Inference (GPU-only):} GPTQ \cite{gptq} and llama.cpp \cite{llamacpp} quantization. These methods run the quantized model exclusively on the GPU without any error compensation.
    \item \textbf{Full-precision Offloading:} vLLM (v0.20.0) \cite{vllm}, llama.cpp (v0.3.20) \cite{llamacpp}, and Hugging Face Accelerate (v1.13.0) \cite{accelerate}. These methods offload part or all of the model parameters to CPU memory to run in full precision.
\end{enumerate}

\textbf{Datasets:} We selected ARC-Easy, ARC-Challenge \cite{arc}, MathQA \cite{mathqa}, MMLU \cite{mmlu}, WikiText2, and C4 \cite{c4} as the evaluation benchmarks.

\subsection{End-to-End Performance}

\begin{figure}
    \centering
    \includegraphics[width=1\textwidth]{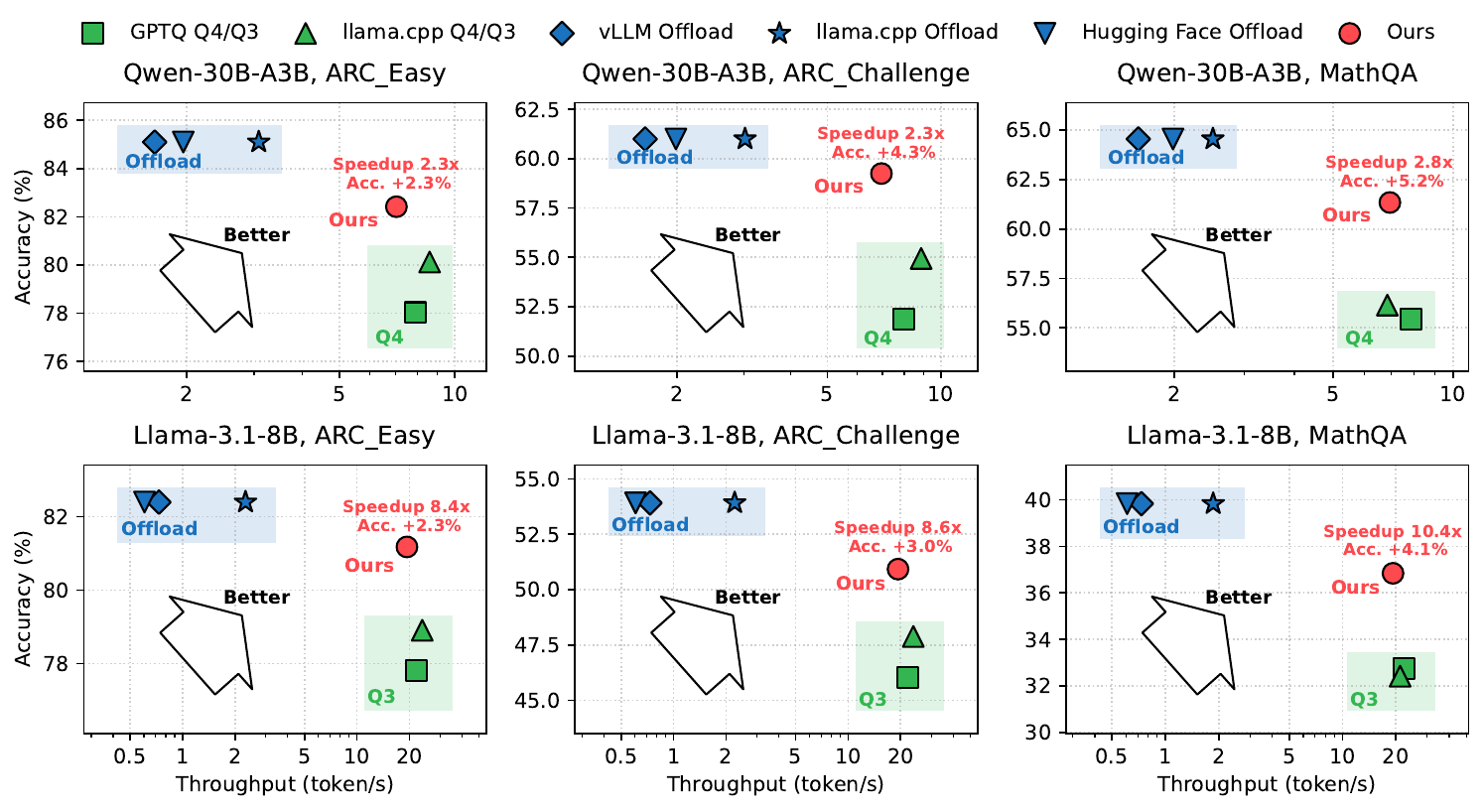}
    \caption{End-to-end throughput and accuracy comparison of \methodname across models and datasets. The green section represents high throughput region with significant accuracy loss; the blue section represents high accuracy area with low inference efficiency.}
    \label{fig:overall}
\end{figure}

In \figref{fig:overall}, we compare the throughput and accuracy of \methodname against baseline methods across various datasets. In tests with the Qwen-30B-A3B model, \methodname achieved significant accuracy improvements of $+2.3\%$, $+4.3\%$, and $+5.2\%$ on the ARC-Easy, ARC-Challenge, and MathQA tasks, respectively, compared to the best quantized baseline. Simultaneously, the throughput of \methodname reached $2.3\times$ to $2.8\times$ that of the offloading schemes. In the memory-constrained scenario using the more aggressive 3-bit quantization for Llama-3.1-8B, \methodname achieved up to a $10.4\times$ throughput improvement on MathQA, and $8.4\times$ and $8.6\times$ improvements on ARC-Easy and ARC-Challenge, respectively, while outperforming the quantized model's accuracy by $2.3\%$ to $4.1\%$.
These results demonstrate that \methodname successfully enhance model accuracy while maintaining high throughput.

\subsection{Fine-grained Analysis}

In this section, we delve into the operational efficiency of \methodname under different workloads and showcase its precision across multiple NLP tasks.

\begin{figure}[h]
    \centering
    \includegraphics[width=1\textwidth]{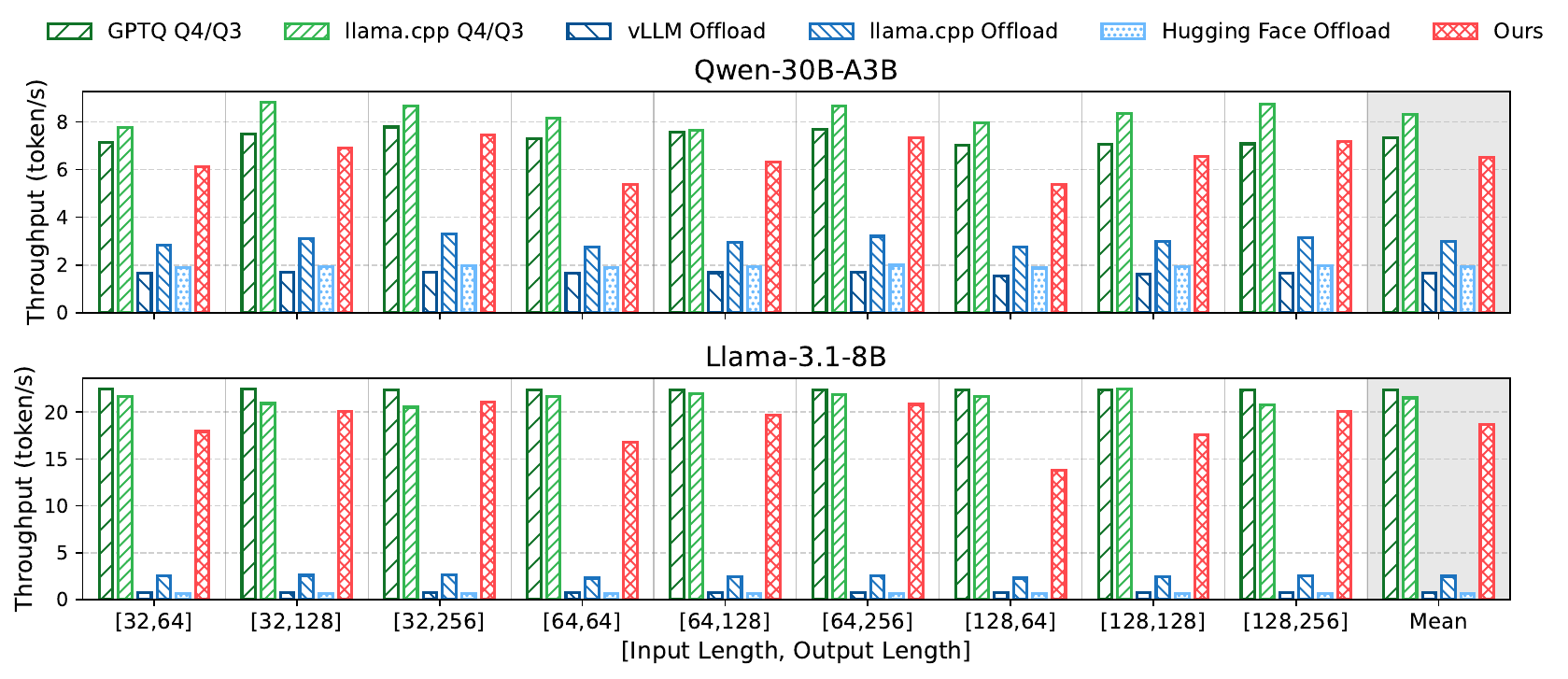}
    \caption{End-to-end performance comparison for various input and output length combinations.}
    \label{fig:throughput}
\end{figure}

\textbf{Throughput across different sequence lengths:} \figref{fig:throughput} illustrates the throughput performance for various input and output length combinations. The results show that \methodname exhibits a clear advantage over full-precision offloading: for Qwen-30B-A3B, throughput improved by $1.9\times$ to $2.2\times$; for Llama-3.1-8B, it improved by $6.0\times$ to $8.2\times$. Our method's advantage is more pronounced in scenarios with longer output lengths, as the CPU's limited parallel processing capability for multiple tokens makes prompt processing significantly less efficient than token generation.

\textbf{Accuracy and Perplexity:} \tabref{accuracy} details the comparison between \methodname and other methods regarding language modeling perplexity and downstream task accuracy. Compared to quantized models, \methodname significantly improves downstream accuracy: $2.2\%$ to $5.2\%$ for Qwen-30B-A3B and $1.7\%$ to $3.6\%$ for Llama-3.1-8B, approaching full-precision levels. For language modeling, \methodname reduces perplexity on WikiText2 and C4 by $0.6$ to $5.1$ for Qwen-30B-A3B and $2.7$ to $2.8$ for Llama-3.1-8B.

\begin{table}[h]
\centering
\caption{Perplexity and accuracy comparison across datasets. Best results are marked in bold.}
\label{accuracy}
\resizebox{\textwidth}{!}{%
\begin{tabular}{llcccccc}
\toprule
\textbf{Model} & \textbf{Method} & \textbf{WikiText2 ($\downarrow$)} & \textbf{C4 ($\downarrow$)} & \textbf{ARC-e ($\uparrow$)} & \textbf{ARC-c ($\uparrow$)} & \textbf{MathQA ($\uparrow$)} & \textbf{MMLU ($\uparrow$)} \\
\midrule
\multirow{4}{*}{Qwen-30B-A3B} & Full Precision & 8.80 & 18.32 & 85.10\% & 61.00\% & 64.55\% & 81.93\% \\
\cmidrule{2-8}
 & GPTQ INT4 & 11.32 & 24.39 & 78.03\% & 51.87\% & 55.44\% & 76.40\% \\
 & Llama.cpp INT4 & 10.03 & 24.57 & 80.13\% & 54.95\% & 56.15\% & 76.85\% \\
 & \textbf{\methodname} & \textbf{9.43} & \textbf{19.51} & \textbf{82.42\%} & \textbf{59.25\%} & \textbf{61.34\%} & \textbf{79.92\%} \\ \midrule
\multirow{4}{*}{Llama-3.1-8B} & Full Precision & 8.81 & 20.15 & 82.40\% & 53.92\% & 39.83\% & 67.90\% \\
\cmidrule{2-8}
 & GPTQ INT3 & 12.696 & 26.53 & 77.81\% & 46.02\% & 32.76\% & 64.19\% \\
 & Llama.cpp INT3 & 12.050 & 24.89 & 78.91\% & 48.89\% & 33.24\% & 65.08\% \\
 & \textbf{\methodname} & \textbf{9.32} & \textbf{22.13} & \textbf{81.18\%} & \textbf{50.93\%} & \textbf{36.84\%} & \textbf{66.74\%} \\ \bottomrule
\end{tabular}%
}
\end{table}

\subsection{Ablation Studies}

We conducted two sets of ablation experiments to verify the core design of \methodname.

\begin{wrapfigure}{r}{0.5\textwidth}
    \centering
    \includegraphics[width=0.5\textwidth]{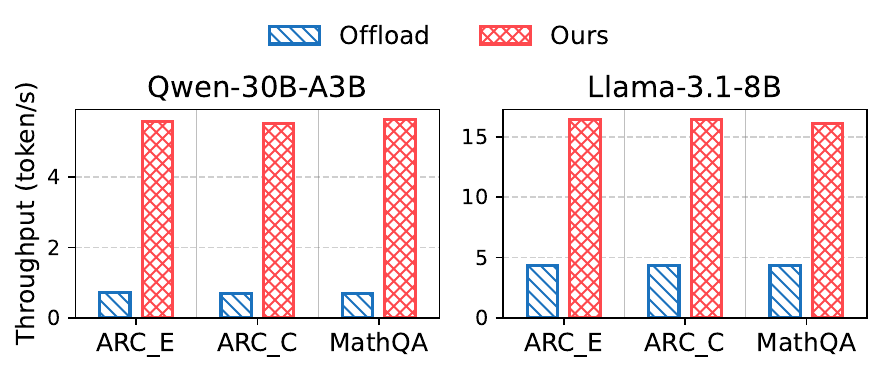}
    \caption{Throughput of \methodname w/ and w/o heterogeneous pipeline. The baseline is a naive offloading scheme where all compensation weights are offloaded to CPU DRAM and moved to GPU only when accessed.}
    \label{fig:abl-pipeline}
\end{wrapfigure}

\textbf{Pipeline Ablation:} To verify the contribution of heterogeneous parallel computing, we compared \methodname with a naive compensation parameter offloading scheme.
\figref{fig:abl-pipeline} shows that \methodname improves throughput by $3.7\times$ to $8.0\times$ over the naive version. \methodname avoids slow parameter loading by transmitting only intermediate activation.

\begin{figure}[h]
    \centering
    \includegraphics[width=0.9\textwidth]{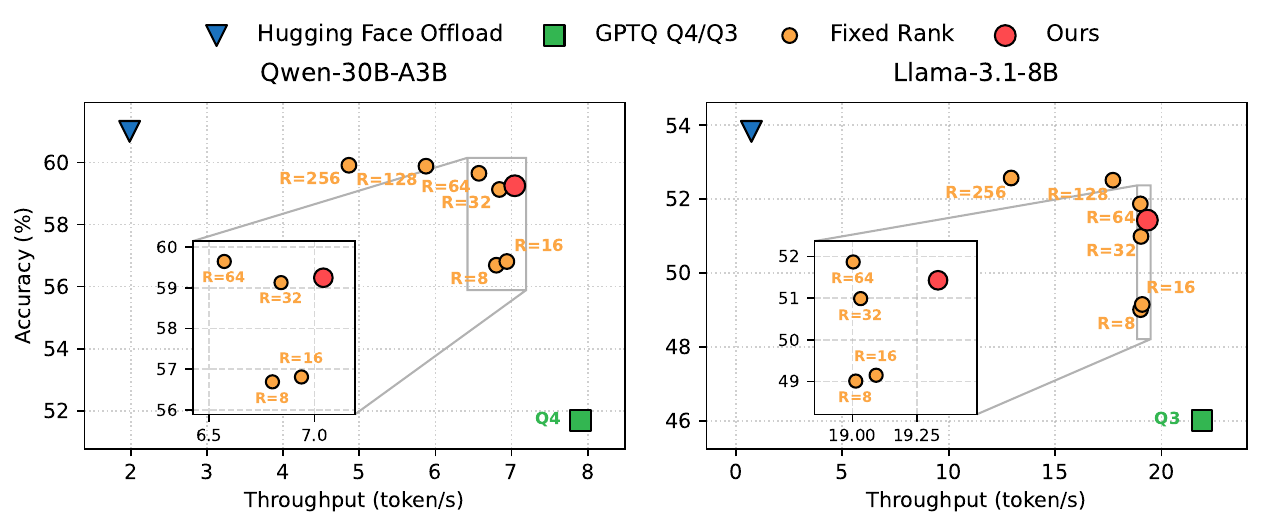}
    \caption{Throughput and accuracy comparison of \methodname w/ and w/o dynamic rank allocator. $R=k$ refers to a fixed-rank allocation scheme where all matrices are assigned with rank-$k$ compensation.}
    \label{fig:abl-throughput}
\end{figure}

\textbf{Parameter Allocation Ablation:} We compared \methodname against fixed-rank compensation schemes. \figref{fig:abl-throughput} shows that \methodname's throughput exceeds all fixed-rank schemes while matching the accuracy of a rank-32 compensation. The throughput gains stem from dynamically skipping compensation for certain matrices, reducing communication and kernel loading overhead. Other experiments yielded similar results, which are listed in \appendixref{sec:appendix-experiment}.

%% file: body/6-conclusion.tex
\section{Conclusion}
\label{sec:conclusion}

In this paper, we introduced \methodname, an efficient heterogeneous inference system that addresses the memory bottlenecks of LLMs on consumer-grade hardware. By integrating quantized base matrices on GPUs with low-rank residual compensation on CPUs, our method achieves high throughput, high accuracy inference. Comprehensive evaluations demonstrate that \methodname achieves near full-precision accuracy while maintaining a peak speedup of x10.4 compared to full precision models. Future work will extend \methodname to broader model architectures to further enhance deployment efficiency.

%% file: body/appedix.tex
\appendix

\section{Implementation Details and Algorithm Workflow}
\label{sec:appendix-pipeline}

\subsection{Implementation Details}

\subsubsection{Dual-Process Execution Architecture}

\methodname is implemented using a persistent dual-process architecture to enable efficient heterogeneous execution. At the beginning of inference, two long-lived processes are initialized:

\begin{itemize}[leftmargin=2em, topsep=-2pt, parsep=0pt]
    \item \textbf{GPU main process.} This process resides on the GPU and is responsible for executing the quantized backbone model, as well as orchestrating the overall inference workflow.
    \item \textbf{CPU worker process.} This process is forked from the main process but does not inherit the loaded model parameters. Instead, it independently loads the LoRA-style compensation parameters and is dedicated to executing residual compensation tasks when triggered.
\end{itemize}

The two processes communicate via shared memory using Python's \texttt{multiprocessing} primitives, including shared variables and event-based synchronization. This design avoids redundant data copies and minimizes communication overhead.

\subsubsection{Asynchronous Communication and Execution}

From the perspective of the GPU process, inference proceeds in a sequence of task blocks. When reaching a block that requires compensation, the GPU process performs the following steps:

\begin{enumerate}[leftmargin=2em, topsep=-2pt, parsep=0pt]
    \item Writes the current activation tensor into shared memory.
    \item Signals the CPU process via an event to start compensation computation.
    \item Immediately continues executing the quantized backbone operators.
\end{enumerate}

After completing the corresponding backbone computation, the GPU process waits for the CPU process to finish, retrieves the compensated outputs from shared memory, and merges them with the current activations before proceeding.

From the CPU process perspective, it acts as a lightweight task executor. It continuously waits for task signals from the GPU process. Upon receiving a signal, it:

\begin{enumerate}[leftmargin=2em, topsep=-2pt, parsep=0pt]
    \item Reads the activation tensor from shared memory.
    \item Determines the current execution context based on its locally maintained state.
    \item Selects the corresponding subset of LoRA parameters according to the dynamic rank allocation results.
    \item Computes the compensation outputs sequentially.
    \item Writes the results back to shared memory and signals completion.
\end{enumerate}

To reduce communication overhead, the CPU process maintains its own synchronized execution state (e.g., current layer and projection index), eliminating the need for the GPU process to repeatedly transmit meta-information. As a result, only activation tensors are exchanged during runtime.

\subsubsection{Computation and Communication Co-Design}
\label{sec:appendix-window}

Naively, LoRA-style compensation operates at the granularity of individual weight matrices. A straightforward implementation would require launching a compensation task for each projection independently, leading to:

\begin{itemize}[leftmargin=2em, topsep=-2pt, parsep=0pt]
    \item Frequent activation transfers and synchronization events, introducing significant communication latency.
    \item Fine-grained CPU workloads with poor utilization.
    \item Strict per-matrix synchronization between GPU and CPU, preventing effective overlap.
\end{itemize}

Moreover, if the compensation computation of a particular projection exceeds the execution time of its corresponding quantized operator, the excess latency directly impacts the end-to-end inference time. This design also prevents different projections from flexibly sharing computation windows based on their importance.

To address these limitations, \methodname introduces computation and communication fusion, which groups multiple operators into shared compensation windows.

\paragraph{Attention Blocks.}
The computation graph of attention can be expressed as:
\[
X \rightarrow W_Q(X), W_K(X), W_V(X) \rightarrow \mathrm{Attn}\left(\frac{QK^\top}{\sqrt{d}}\right)V \rightarrow W_O(\cdot).
\]
Since the $Q$, $K$, and $V$ projections are independent, \methodname fuses them into a single compensation window, allowing them to share CPU computation and communication. The output projection $W_O$ is processed in a separate window due to its dependency on the attention result.

\paragraph{Dense Feed-Forward Networks (FFN).}
The FFN computation is:
\[
X \rightarrow W_{\text{up}}(X), W_{\text{gate}}(X) \rightarrow h = \sigma(W_{\text{gate}}(X)) \odot W_{\text{up}}(X) \rightarrow W_{\text{down}}(h).
\]
Here, $W_{\text{up}}$ and $W_{\text{gate}}$ are independent and thus fused into a shared compensation window, while $W_{\text{down}}$ is assigned a separate window.

\paragraph{MoE Layer.}
For MoE layers, the computation is:
\[
X \rightarrow \mathrm{Router}(X) \rightarrow 
\Big( W_{\text{up}}(X), W_{\text{gate}}(X) \rightarrow h \rightarrow W_{\text{down}}(h) \Big) \times k \rightarrow \sum_i w_i E_i(X).
\]
In addition to projection-level independence, MoE introduces independence across experts. \methodname further reorganizes execution by grouping operators across all activated experts: All experts' $W_{\text{up}}$ and $W_{\text{gate}}$ share a single compensation window, and all experts' $W_{\text{down}}$ share another compensation window. This design enables cross-expert sharing of CPU computation windows, significantly improving resource utilization.

\subsubsection{Computation Pipeline Diagram}

\begin{figure}[hb]
    \centering
    \includegraphics[width=\textwidth]{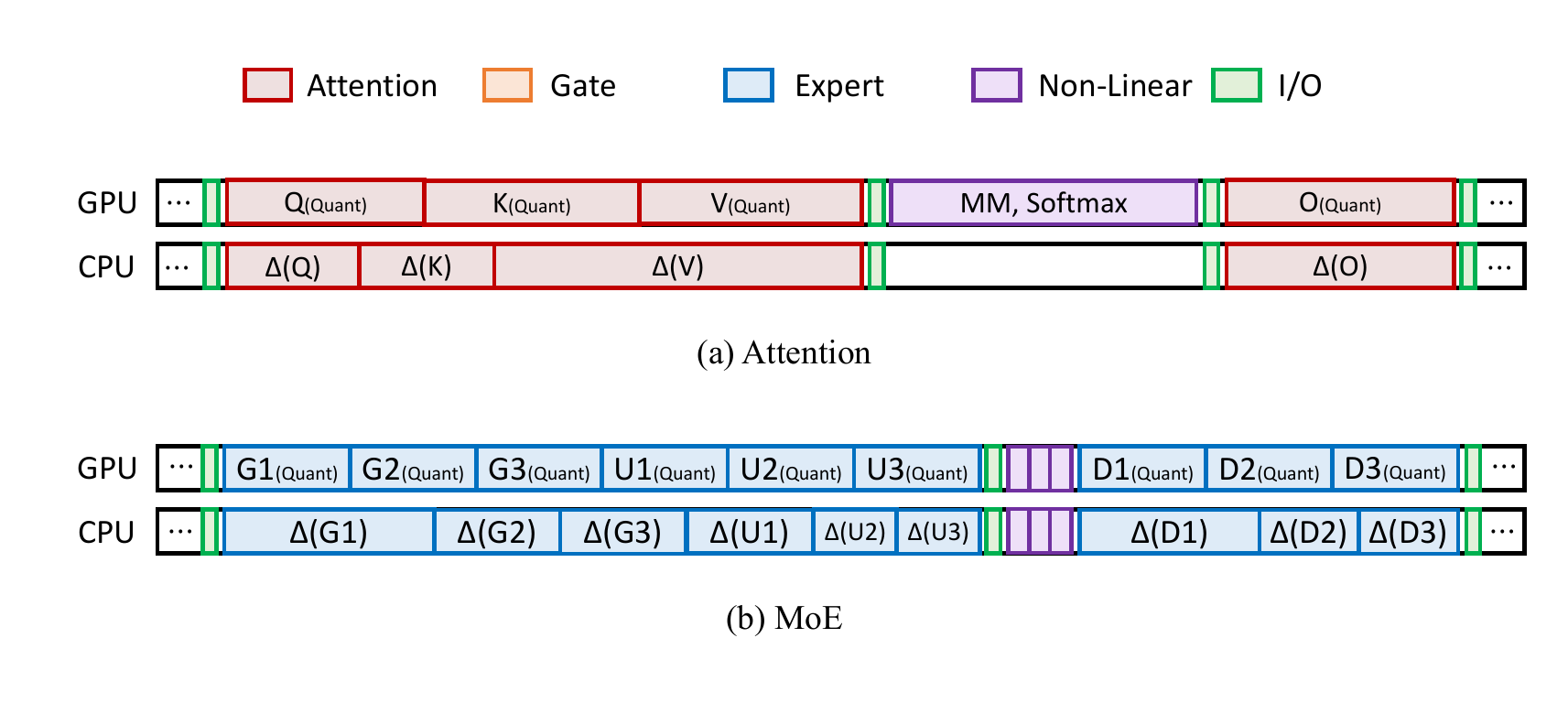}
    \caption{Detailed illustration of the heterogeneous execution pipeline. The purple blocks in (b) denote the non-linear activation functions for each expert. To reduce communication overhead, these activations are recomputed on the CPU rather than transferred from the GPU, leveraging their negligible computational cost.}
    \label{fig:pipeline-detail}
\end{figure}

We illustrate the compensation window sharing and execution pipeline in \figref{fig:pipeline-detail}. By fusing communication and computation across operators and experts, \methodname establishes a structured heterogeneous execution pipeline that maximizes overlap between GPU and CPU.

Specifically, this design reduces synchronization frequency and communication overhead,enables effective overlap between GPU and CPU execution,and allows multiple matrices to share a unified compensation budget within the same execution window. Importantly, this pipeline organization maximizes the flexibility of dynamic rank allocation, allowing compensation resources to be adaptively distributed across matrices according to their importance while maintaining high system efficiency.

\subsection{End-to-End Algorithm Workflow}

We summarize the end-to-end execution of \methodname using two cooperating processes: a GPU process that executes the quantized backbone and controls the inference flow, and a CPU process that performs asynchronous LoRA-based compensation.

\algref{alg:gpu_process} describes the GPU-side execution, including activation dispatch, backbone computation, and result merging. \algref{alg:cpu_process} describes the CPU worker, which continuously fetches tasks, applies dynamic rank allocation, and computes compensation outputs. Together, these two processes form an asynchronous pipeline that overlaps computation and communication.

\begin{algorithm}[H]
\caption{GPU Process: Quantized Backbone with Asynchronous Compensation}
\label{alg:gpu_process}
\begin{algorithmic}[1]
\STATE \textbf{Input:} Input tokens $X$
\STATE \textbf{Shared memory:} $\mathcal{M}$
\STATE \textbf{Events:} $\texttt{cpu\_start}$, $\texttt{cpu\_done}$

\STATE Initialize model state and layer index $\ell \leftarrow 1$

\FOR{each generation step}
\FOR{each layer $\ell$}
    \STATE Identify computation blocks (attention / FFN / MoE)

    \FOR{each compensation window $w$ in layer $\ell$}
        \STATE Extract activation $A_w$
        \STATE Write $A_w$ to shared memory $\mathcal{M}$
        \STATE Set $\texttt{cpu\_start}$ event
        \STATE Compute $\hat{Y}_w$ using quantized operators on GPU
        \STATE Wait until $\texttt{cpu\_done}$ is triggered
        \STATE Read $\Delta Y_w$ from shared memory $\mathcal{M}$
        \STATE $Y_w \leftarrow \hat{Y}_w + \Delta Y_w$
        \STATE Clear $\texttt{cpu\_done}$ event
    \ENDFOR

    \STATE Update layer state $\ell \leftarrow \ell + 1$
\ENDFOR
\ENDFOR

\STATE \textbf{Return:} Final output
\end{algorithmic}
\end{algorithm}

\begin{algorithm}[H]
\caption{CPU Process: LoRA-Based Compensation Worker}
\label{alg:cpu_process}
\begin{algorithmic}[1]
\STATE \textbf{Shared memory:} $\mathcal{M}$
\STATE \textbf{Events:} $\texttt{cpu\_start}$, $\texttt{cpu\_done}$

\STATE Load LoRA compensation parameters
\STATE Initialize local execution state (layer index, operator type)

\WHILE{inference is running}
    \STATE Wait for $\texttt{cpu\_start}$ event
    \STATE Read activation $A_w$ from shared memory $\mathcal{M}$
    \STATE Identify current layer $\ell$ and window type $w$
    \STATE Retrieve allocated ranks $\{r_i\}$ for matrices in window $w$
    \STATE $\Delta Y_w \leftarrow 0$
    \FOR{each matrix $i$ in window $w$}
        \STATE Select LoRA parameters $(A_i, B_i)$ with rank $r_i$
        \STATE $\Delta Y_w \leftarrow \Delta Y_w + (A_w A_i) B_i$
    \ENDFOR
    \STATE Write $\Delta Y_w$ to shared memory $\mathcal{M}$
    \STATE Update local execution state
    \STATE Clear $\texttt{cpu\_start}$ event
    \STATE Set $\texttt{cpu\_done}$ event
\ENDWHILE
\end{algorithmic}
\end{algorithm}

\section{Theoretical Analysis of Dynamic Rank Allocation}
\label{sec:appendix-algorithm}

\subsection{Mathematical Formulation of the Dynamic Rank Allocator}
The sensitivity-aware dynamic rank allocator in \methodname integrates three complementary factors to determine the allocation priority for each weight matrix: 
(i) \emph{intrinsic recoverability}, captured by the singular value salience score $\mathcal{V}_i$; 
(ii) \emph{output sensitivity}, characterized by matrix- and layer-level sensitivity scores $\mathcal{S}_i$ and $\mathcal{S}_\ell$; 
and (iii) \emph{expert activation}, modeled by the expert-level score $\mathcal{G}_{i,e}$ for MoE architectures. 
We describe each component in detail below.

\paragraph{Singular Value Salience Score $\mathcal{V}_i$.}

We measure the intrinsic recoverability of a quantization residual matrix via the concentration of its singular value spectrum. 
For a given residual matrix $\Delta W_i \in \mathbb{R}^{d_i \times k_i}$, we perform singular value decomposition:
\[
\Delta W_i = U_i \Sigma_i V_i^\top, \quad 
\Sigma_i = \mathrm{diag}(\sigma_{i,1}, \sigma_{i,2}, \ldots, \sigma_{i,n_i}),
\]
where $\sigma_{i,1} \ge \sigma_{i,2} \ge \cdots \ge \sigma_{i,n_i} \ge 0$.

We first normalize the singular values:
\[
\hat{\sigma}_{i,j} = \frac{\sigma_{i,j}}{\sigma_{i,1}}.
\]

To identify the salient components, we compute the second-order discrete difference:
\[
k_{i,j} = \hat{\sigma}_{i,j-1} - 2\hat{\sigma}_{i,j} + \hat{\sigma}_{i,j+1}.
\]

Let 
\[
r_i = \arg\max_j k_{i,j}.
\]
If $\max_j k_{i,j} > \tau$, where $\tau$ is a predefined threshold (set to $0.01$ in our implementation), we partition the singular values into:
\[
S_i = \{1, \ldots, r_i\}, \quad
R_i = \{r_i + 1, \ldots, n_i\}.
\]
Otherwise, we set $S_i = \emptyset$ and $R_i = \{1, \ldots, n_i\}$.

We then define the salience ratio:
\[
\phi_i =
\begin{cases}
\displaystyle
\frac{\frac{1}{|S_i|} \sum_{j \in S_i} \sigma_{i,j}}
{\frac{1}{|R_i|} \sum_{j \in R_i} \sigma_{i,j}}, 
& S_i \neq \emptyset, \\[10pt]
1, & S_i = \emptyset.
\end{cases}
\]

Finally, we normalize $\phi_i$ within each compensation window $\mathcal{W}$ (as defined in \appendixref{sec:appendix-window}) to obtain:
\[
\mathcal{V}_i = \frac{\phi_i}{\sum_{j \in \mathcal{W}} \phi_j}.
\]

\paragraph{Matrix- and Layer-Level Sensitivity $\mathcal{S}_i, \mathcal{S}_\ell$.}

To quantify the impact of quantization on model outputs, we measure the divergence between the output distributions before and after replacing a matrix with its quantized version.

For a given matrix $i$, we define:
\[
D_i = D_{\mathrm{KL}} \big( P(Y | X) \,\|\, Q_i(Y | X) \big),
\]
where $P(Y | X)$ denotes the output distribution of the full-precision model, and $Q_i(Y | X)$ denotes the output distribution after replacing matrix $i$ with its quantized counterpart. The KL divergence is computed as:
\[
D_{\mathrm{KL}}(P \| Q) = \sum_{y} P(y | X) \log \frac{P(y | X)}{Q(y | X)}.
\]

We normalize matrix-level sensitivity within each compensation window:
\[
\mathcal{S}_i = \frac{D_i}{\sum_{j \in \mathcal{W}} D_j}.
\]

For layer-level sensitivity, we replace all matrices in layer $\ell$ with their quantized versions and compute:
\[
D_\ell = D_{\mathrm{KL}} \big( P(Y | X) \,\|\, Q_\ell(Y | X) \big).
\]

To emphasize the most critical layers under limited computation budgets, we select the top-$K$ layers with the largest $D_\ell$, denoted as $\mathcal{T}$. The layer-level sensitivity score is defined as:
\[
\mathcal{S}_\ell =
\begin{cases}
1, & \ell \in \mathcal{T}, \\[6pt]
\displaystyle
\frac{D_\ell}{\min_{m \in \mathcal{T}} D_m}, & \ell \notin \mathcal{T}.
\end{cases}
\]

This formulation ensures that highly sensitive layers receive uniformly maximal priority, while less sensitive layers are proportionally down-weighted.

\paragraph{Expert Activation Score $\mathcal{G}_{i,e}$.}

For MoE architectures, the contribution of each expert is modulated by routing probabilities. 
Let $\mathcal{A}$ denote the set of activated experts, and $g_e$ be the routing weight for expert $e \in \mathcal{A}$, satisfying:
\[
\sum_{e \in \mathcal{A}} g_e = 1.
\]

Given that $k = |\mathcal{A}|$ experts are activated, we define:
\[
\mathcal{G}_{i,e} = k \cdot g_e.
\]

This scaling redistributes the normalized routing probabilities across the activated experts, ensuring that the total contribution is consistent with the number of active experts.

\paragraph{Rank Allocation Strategy.}

Given the unified priority score, we allocate the compensation rank proportionally under a hardware-constrained budget. 
For matrix $i$ (and expert $e$ in MoE models), we first define the normalized priority:
\[
\mathcal{P}_{i,e} = \mathcal{G}_{i,e} \cdot \mathrm{Norm}_{\mathcal{W}}(\mathcal{V}_i \cdot \mathcal{S}_i) \cdot \mathcal{S}_\ell,
\]
where $\mathrm{Norm}_{\mathcal{W}}(\cdot)$ denotes normalization within the corresponding compensation window $\mathcal{W}$.

Let $r_{\mathrm{std}}$ denote the reference rank determined by the hardware constraint. 
The initial continuous rank allocation is given by:
\[
\tilde{r}_{i,e} = \mathcal{P}_{i,e} \cdot r_{\mathrm{std}}.
\]

\paragraph{Two-Stage Singular Value Allocation.}

Recall that the singular values of each residual matrix are partitioned into a salient set $S_i$ and a residual set $R_i$. 
To maximize reconstruction effectiveness, we adopt a two-stage allocation strategy:

\begin{itemize}[leftmargin=2em, topsep=-2pt, parsep=0pt]
    \item \textbf{Stage 1 (Salient allocation).} 
    Ranks are first assigned proportionally across matrices using only the singular values in $S_i$. 
    That is, each matrix preferentially selects components from its salient subspace according to $\tilde{r}_{i,e}$.

    \item \textbf{Stage 2 (Residual allocation).} 
    If the total allocated rank exceeds the total size of all salient sets, the remaining budget is further distributed proportionally over the residual sets $R_i$.
\end{itemize}

This design ensures that the most informative low-rank components are utilized first, improving the efficiency of limited rank budgets.

\paragraph{Discrete Rank Alignment.}

To facilitate efficient parallel execution on the CPU, we restrict the final rank values to a discrete set aligned with hardware-friendly sizes. 
Specifically, we constrain:
\[
r_{i,e} \in \{0\} \cup \{2^k \mid k \ge k_0\},
\]
where $k_0$ is chosen such that the minimum non-zero rank (e.g., $2^{k_0}=8$ in our implementation) balances computational granularity and overhead.

The final rank is obtained by rounding $\tilde{r}_{i,e}$ to the nearest admissible value:
\[
r_{i,e} = \mathrm{Align}(\tilde{r}_{i,e}),
\]
where $\mathrm{Align}(\cdot)$ denotes projection onto the nearest allowed rank level.

This alignment improves SIMD efficiency and memory access patterns on the CPU, while introducing negligible distortion to the proportional allocation.

\subsection{Optimality Analysis of Allocation Strategy}
\label{sec:appendix-proof}

In this section, we analyze the optimality of the proposed dynamic rank allocation strategy. 
We show that, under mild assumptions, allocating ranks proportionally to the priority score 
is an optimal solution to a constrained resource allocation problem.

\paragraph{Problem Formulation.}

The goal of the proposed inference system is to maximize the overall compensation effect under pipeline constraints. Let $Y$ denote the model output, and $\Delta Y$ denote the model compensation effect. Thus, the allocation problem can be formulated as:
\[
\max_{\{r_i\}} \sum_{i \in \mathcal{W}} \Delta Y
\quad \text{s.t.} \quad
\sum_{i\in\mathcal{W}}T_{\mathrm{CPU}}(r_i) + T_{\mathrm{comm},\mathcal{W}} \le T_{\mathrm{GPU},\mathcal{W}}, \quad r_i \ge 0.
\]

The constraints are imposed by the heterogeneous execution pipeline in \appendixref{sec:appendix-window}, where for each compensation window $\mathcal{W}$, the CPU computation and communication must be fully hidden behind GPU execution.

We start our analysis from the fundamental constraint imposed by the heterogeneous execution pipeline. Due to CPU's low computational parallelism, the CPU computation latency is approximately linear in the rank:
\[
T_{\mathrm{CPU}}(r_i) \approx k \cdot r_i,
\]
where $k$ is a hardware-dependent constant. We define the standard rank $r_{\mathrm{std}}$ as the maximum rank satisfying:
\[
T_{\mathrm{CPU}}(r_{\mathrm{std}}) + T_{\mathrm{comm}} = T_{\mathrm{GPU}}.
\]

Therefore, the pipeline constraint can be rewritten as:
\[
\sum_{i\in\mathcal{W}}T_{\mathrm{CPU}}(r_i) \le T_{\mathrm{CPU}}(r_{\mathrm{std}}).
\]

Since the total CPU workload is the sum of all compensation tasks, we have:
\[
\sum_{i\in\mathcal{W}}T_{\mathrm{CPU}}(r_i)\approx T_{\mathrm{CPU}}\left(\sum_{i\in\mathcal{W}}r_i\right),
\]
which leads to the equivalent constraint:
\[
\sum_{i \in \mathcal{W}} r_i \le r_{\mathrm{std}}, \quad r_i \ge 0.
\]

We then analyze the optimization objectives. Since compensation introduces a small perturbation, we approximate the improvement using a first-order expansion:
\[
\Delta Y \approx \sum_{i \in \mathcal{W}} \frac{\partial Y}{\partial r_i} \, r_i.
\]

Thus, the allocation problem can be formulated as:
\[
\max_{\{r_i\}} \sum_{i \in \mathcal{W}} \frac{\partial Y}{\partial r_i} \, r_i
\quad \text{s.t.} \quad
\sum_{i \in \mathcal{W}} r_i \le r_{\mathrm{std}}, \quad r_i \ge 0.
\]

This formulation reduces the dynamic rank allocation problem to a constrained resource allocation problem within each compensation window.

\paragraph{Optimal Allocation via Greedy Strategy.}

We define marginal gain as
\[
g_i = \frac{\partial Y}{\partial r_i}.
\]
$g_i$ is non-increasing with respect to $r_i$, so the above problem is a classical resource allocation problem. Its optimal solution is to allocate resources in descending order of $g_i$, which is equivalent to a proportional allocation:
\[
r_i \propto g_i.
\]

This result can be obtained via a greedy argument or by observing that the problem reduces to a linear program whose optimal solution lies at an extreme point.

\paragraph{Approximation of $\partial Y / \partial r_i$.}

We now derive an explicit approximation of the marginal gain 
$\frac{\partial Y}{\partial r_i}$ that governs the optimal allocation.

Using the chain rule, we decompose:
\[
\frac{\partial Y}{\partial r_i}
=
\frac{\partial Y}{\partial W_i}
\cdot
\frac{\partial W_i}{\partial r_i},
\]

\textbf{(1) Output Sensitivity Term.} The first term measures how sensitive the model output is to perturbations in $W_i$.

We approximate this quantity using the KL divergence between the original and perturbed outputs. Let $P(Y | X)$ be the original output distribution and $Q_i(Y | X)$ be the distribution after perturbing $W_i$. Then:
\[
D_i = D_{\mathrm{KL}}\big(P(Y | X)\,\|\,Q_i(Y | X)\big).
\]

Using a first-order perturbation approximation, we have:
\[
\left\| \frac{\partial Y}{\partial W_i} \right\|
\;\propto\;
D_i.
\]

After normalization within a compensation window $\mathcal{W}$, this yields:
\[
\left\|\frac{\partial Y}{\partial W_i}\right\|
\;\approx\;
\mathcal{S}_i.
\]

\paragraph{(2) Weight Recoverability Term.}

The second term $\frac{\partial W_i}{\partial r_i}$ quantifies how much of the quantization error 
can be recovered as the rank increases.

Let $\Delta W_i$ denote the quantization residual with singular values $\{\sigma_{i,j}\}$. According to the Eckart--Young--Mirsky theorem, the optimal rank-$r$ approximation satisfies:
\[
\| \Delta W_i - (\Delta W_i)_r \|_F^2 = \sum_{j > r} \sigma_{i,j}^2.
\]

Therefore, the marginal gain from increasing the rank by one is:
\[
\frac{\partial}{\partial r} \| (\Delta W_i)_r \|_F^2 
\;\approx\;
\sigma_{i,r}^2.
\]

This implies:
\[
\left\|\frac{\partial W_i}{\partial r_i}\right\|
\;\propto\;
\sigma_{i,r}.
\]

To obtain a stable and scale-invariant measure, we aggregate this effect using the singular value salience score:
\[
\left\|\frac{\partial W_i}{\partial r_i}\right\|
\;\approx\;
\mathcal{V}_i.
\]

\textbf{(3) MoE Activation Term.} For MoE layer, the output is given by:
\[
Y_\text{MoE} = \sum_{e \in \mathcal{A}} g_e \, E_e(X),
\]
where $g_e$ are routing weights and $\sum_e g_e = 1$.

Let $W_{i,e}$ denote the parameters of matrix $i$ in expert $e$. Then:
\[
\frac{\partial Y}{\partial W_{i,e}}
=
g_e \cdot \frac{\partial E_e(X)}{\partial W_{i,e}}.
\]

Thus, the marginal gain is scaled by:
\[
\left\|\frac{\partial Y}{\partial W_{i,e}}\right\|
\;\propto\;
g_e,
\]
which corresponds to the expert activation score:
\[
\mathcal{G}_{i,e} = k \cdot g_e.
\]

\paragraph{Combined Expression.}

Combining all terms, we obtain:
\[
\frac{\partial Y}{\partial r_{i,e}}
\;\propto\;
\mathcal{V}_i \cdot \mathcal{S}_i \cdot \mathcal{G}_{i,e}.
\]

Substituting this into the optimization problem yields:
\[
r_{i,e} \propto 
\mathcal{V}_i \cdot \mathcal{S}_i \cdot \mathcal{G}_{i,e},
\]
which is exactly the allocation rule adopted in \methodname.

\section{Extensive Experimental Analysis}
\label{sec:appendix-experiment}

In this section, we provide additional experimental analysis to further understand the behavior of \methodname. 
We focus on two aspects: 
(i) the characteristics and robustness of the proposed dynamic rank allocation strategy, and 
(ii) the runtime behavior of the heterogeneous execution pipeline.

\subsection{Rank Allocation Profiles across Model Layers}
\label{sec:appendix_rank_profile}

To better understand the effectiveness of the proposed allocator, we analyze the allocation profiles produced by \methodname and evaluate nearby alternative allocation strategies.

Specifically, starting from the allocation generated by our sensitivity-aware allocator, we construct multiple perturbed allocation schemes by redistributing compensation ranks across different layers and operator groups while keeping the overall rank budget unchanged. 
We then measure both model accuracy and inference throughput for each configuration.

\tabref{tab:appendix-rank} summarizes the results. 
We observe that the allocation produced by \methodname\ consistently achieves a superior balance between accuracy and throughput compared with neighboring allocation strategies. 
In particular, manually shifting compensation ranks toward less sensitive layers or operator groups generally leads to noticeable accuracy degradation while providing limited throughput improvement.

These results validate the core design principle of our allocator: compensation resources should be preferentially assigned to matrices with high marginal utility, rather than distributed uniformly across the model.

Furthermore, we observe that the performance landscape around the optimal allocation is relatively smooth, indicating that the proposed strategy is robust to small perturbations. 
This behavior suggests that the derived allocation scores effectively capture the relative importance of different matrices and layers.

\begin{table}[t]
\centering
\caption{
Accuracy and throughput under different rank allocation strategies. The first row corresponds to the  reference allocation produced by \methodname. Values in parentheses denote the relative change compared with \methodname\ in terms of accuracy and throughput, respectively.
}
\label{tab:appendix-rank}

\begin{tabular}{lcccccc}
\toprule
Rank Distribution 
& Accuracy 
& Throughput (token/s) \\
\midrule

(0, 0, 64, 32, 32, 16, 32)
& \ 59.25\% (+0.0\%)
& \ 6.960 (+0.0\%)
\\

\midrule

(32, 0, 32, 32, 32, 16, 32)
& 55.04\% (-4.2\%)
& 6.935 (+0.7\%)
\\

(0, 32, 32, 32, 32, 16, 32)
& 55.11\% (-4.1\%)
& 6.867 (-0.2\%)
\\

(32, 32, 32, 32, 32, 16, 32)
& 56.88\% (-2.3\%)
& 6.963 (+1.1\%)
\\

(0, 0, 128, 32, 32, 16, 32)
& 59.78\% (+0.5\%)
& 6.574 (-4.4\%)
\\

(64, 0, 64, 32, 32, 16, 32)
& 58.93\% (-0.3\%)
& 6.523 (-5.2\%)
\\

(0, 64, 64, 32, 32, 16, 32)
& 60.02\% (+0.7\%)
& 6.491 (-5.6\%)
\\

(0, 0, 32, 32, 32, 16, 32)
& 55.12\% (-4.1\%)
& 6.890 (+0.0\%)
\\

(16, 0, 16, 32, 32, 16, 32)
& 56.91\% (-2.3\%)
& 6.883 (-0.0\%)
\\

(0, 16, 16, 32, 32, 16, 32)
& 55.04\% (-4.2\%)
& 6.851 (-0.4\%)
\\

(0, 0, 64, 16, 32, 16, 32)
& 54.12\% (-5.1\%)
& 6.963 (+1.1\%)
\\

(0, 0, 64, 8, 32, 16, 32)
& 53.14\% (-6.1\%)
& 6.890 (+0.0\%)
\\

(0, 0, 64, 64, 32, 16, 32)
& 59.71\% (+0.4\%)
& 6.731 (-2.2\%)
\\

(0, 0, 64, 128, 32, 16, 32)
& 59.25\% (+0.0\%)
& 6.436 (-6.4\%)
\\

(0, 0, 64, 32, 16, 32, 32)
& 55.87\% (-3.3\%)
& 6.951 (+0.9\%)
\\

(0, 0, 64, 32, 16, 16, 32)
& 54.19\% (-5.0\%)
& 6.960 (+1.1\%)
\\

(0, 0, 64, 32, 16, 8, 32)
& 54.12\% (-5.1\%)
& 7.024 (+2.0\%)
\\

(0, 0, 64, 32, 64, 16, 32)
& 59.25\% (+0.0\%)
& 6.930 (+0.6\%)
\\

(0, 0, 64, 32, 32, 32, 32)
& 59.71\% (+0.4\%)
& 6.889 (+0.0\%)
\\

(0, 0, 64, 32, 64, 32, 32)
& 60.12\% (+0.8\%)
& 6.848 (-0.5\%)
\\

(0, 0, 64, 32, 32, 16, 16)
& 54.12\% (-5.1\%)
& 6.910 (+0.3\%)
\\

(0, 0, 64, 32, 32, 16, 8)
& 55.87\% (-3.3\%)
& 6.861 (-0.3\%)
\\

(0, 0, 64, 32, 32, 16, 64)
& 59.71\% (+0.4\%)
& 6.849 (-0.4\%)
\\

(0, 0, 64, 32, 32, 16, 128)
& 59.71\% (+0.4\%)
& 6.489 (-5.7\%)
\\

\bottomrule
\end{tabular}
\end{table}

\subsection{Computational Latency Breakdown and Profiling}
\label{sec:appendix_latency}

To analyze the runtime behavior of \methodname, we perform detailed profiling of the heterogeneous execution pipeline during autoregressive decoding.

We first measure the average per-token and per-window latency breakdown across the entire inference process, including:
(i) total CPU compensation time,
(ii) total GPU backbone computation time,
(iii) CPU--GPU communication overhead, and
(iv) effective overlap time between CPU and GPU execution.
The results are summarized in \tabref{tab:appendix-breakdown}. 

\begin{table}[t]
\centering
\caption{
Latency breakdown during autoregressive decoding.
}
\label{tab:appendix-breakdown}

\begin{tabular}{lccccc}
\toprule
Type & Total (ms) & CPU (ms) & GPU (ms) & Comm. (ms) & Overlap (ms) \\
\midrule
Iteration & 128.33 & 67.99 & 87.04 & 3.10 & 57.54 \\
\midrule
Attention QKV & 13.15 & 2.74 & 3.05 & 1.01 & 2.39 \\
Attention O & 4.69 & 1.22 & 1.36 & 0.33 & 1.06 \\
FFN/MoE Gate/Up & 68.07 & 66.39 & 57.28 & 1.08 & 56.67 \\
FFN/MoE Down & 32.13 & 31.22 & 25.35 & 0.68 & 25.12 \\
\bottomrule
\end{tabular}
\end{table}

\vspace{0.5em}

We observe that a substantial fraction of the CPU-side compensation latency is successfully hidden behind GPU execution, validating the effectiveness of the asynchronous pipeline design.

Moreover, the communication overhead accounts for only a small portion of the total runtime, demonstrating that the proposed window-level communication fusion effectively reduces synchronization and data transfer costs.

From per-window breakdown experiment, we observe that different operator groups exhibit distinct latency characteristics. 
In particular, attention projection windows generally achieve higher overlap efficiency due to their larger GPU execution regions, while certain FFN or MoE windows become more sensitive to CPU-side workload imbalance.

These observations further motivate the importance of dynamic rank allocation and window-level scheduling in maximizing hardware utilization.

Finally, we observe that the overlap ratio remains consistently high across most compensation windows, indicating that the proposed pipeline organization successfully converts otherwise exposed CPU computation into effectively hidden latency. 
This behavior is critical for maintaining high throughput while enabling adaptive low-rank compensation.

In \tabref{tab:appendix-breakdown-detail}, we present the detailed profiling results.

\begin{table}[H]
\centering
\caption{
Detailed latency breakdown by layer.
}
\label{tab:appendix-breakdown-detail}

\resizebox{\linewidth}{!}{
\begin{tabular}{llccccc}
\toprule
Layer & Window & Total (ms) & CPU (ms) & GPU (ms) & Comm. (ms) & Overlap (ms) \\
\midrule
layer0 & Attention QKV & 0.37 & 0.29 & 0.30 & 0.06 & 0.25 \\
layer0 & Attention O & 0.12 & 0.08 & 0.09 & 0.03 & 0.07 \\
layer0 & FFN/MoE Gate/Up & 1.54 & 0.22 & 0.24 & 1.31 & 0.20 \\
layer0 & FFN/MoE Down & 0.75 & 0.18 & 0.18 & 0.57 & 0.16 \\
\midrule
layer4 & Attention QKV & 0.27 & 0.21 & 0.21 & 0.06 & 0.18 \\
layer4 & Attention O & 0.10 & 0.06 & 0.07 & 0.03 & 0.05 \\
layer4 & FFN/MoE Gate/Up & 1.38 & 0.20 & 0.22 & 1.15 & 0.18 \\
layer4 & FFN/MoE Down & 0.65 & 0.14 & 0.14 & 0.51 & 0.12 \\
\midrule
layer8 & Attention QKV & 0.25 & 0.17 & 0.19 & 0.06 & 0.15 \\
layer8 & Attention O & 0.09 & 0.06 & 0.07 & 0.03 & 0.06 \\
layer8 & FFN/MoE Gate/Up & 1.33 & 0.19 & 0.20 & 1.13 & 0.17 \\
layer8 & FFN/MoE Down & 0.63 & 0.12 & 0.12 & 0.51 & 0.11 \\
\midrule
layer12 & Attention QKV & 0.26 & 0.20 & 0.20 & 0.06 & 0.17 \\
layer12 & Attention O & 0.09 & 0.06 & 0.07 & 0.03 & 0.06 \\
layer12 & FFN/MoE Gate/Up & 1.35 & 0.22 & 0.23 & 1.12 & 0.20 \\
layer12 & FFN/MoE Down & 0.65 & 0.14 & 0.15 & 0.50 & 0.13 \\
\midrule
layer16 & Attention QKV & 0.28 & 0.20 & 0.21 & 0.07 & 0.18 \\
layer16 & Attention O & 0.10 & 0.07 & 0.07 & 0.03 & 0.06 \\
layer16 & FFN/MoE Gate/Up & 1.38 & 0.11 & 0.11 & 1.27 & 0.10 \\
layer16 & FFN/MoE Down & 0.67 & 0.10 & 0.11 & 0.56 & 0.09 \\
\midrule
layer20 & Attention QKV & 0.25 & 0.18 & 0.18 & 0.06 & 0.16 \\
layer20 & Attention O & 0.09 & 0.06 & 0.07 & 0.03 & 0.05 \\
layer20 & FFN/MoE Gate/Up & 1.34 & 0.15 & 0.16 & 1.18 & 0.13 \\
layer20 & FFN/MoE Down & 0.64 & 0.10 & 0.11 & 0.54 & 0.09 \\
\midrule
layer24 & Attention QKV & 0.25 & 0.17 & 0.18 & 0.06 & 0.15 \\
layer24 & Attention O & 0.09 & 0.06 & 0.06 & 0.03 & 0.05 \\
layer24 & FFN/MoE Gate/Up & 1.31 & 0.13 & 0.14 & 1.18 & 0.11 \\
layer24 & FFN/MoE Down & 0.78 & 0.27 & 0.27 & 0.51 & 0.24 \\
\midrule
layer28 & Attention QKV & 0.27 & 0.19 & 0.20 & 0.07 & 0.17 \\
layer28 & Attention O & 0.10 & 0.07 & 0.07 & 0.03 & 0.06 \\
layer28 & FFN/MoE Gate/Up & 1.46 & 0.28 & 0.29 & 1.17 & 0.24 \\
layer28 & FFN/MoE Down & 0.71 & 0.18 & 0.18 & 0.53 & 0.16 \\
\midrule
layer32 & Attention QKV & 0.27 & 0.20 & 0.21 & 0.06 & 0.17 \\
layer32 & Attention O & 0.10 & 0.07 & 0.07 & 0.03 & 0.06 \\
layer32 & FFN/MoE Gate/Up & 1.33 & 0.17 & 0.17 & 1.16 & 0.15 \\
layer32 & FFN/MoE Down & 0.64 & 0.12 & 0.14 & 0.51 & 0.11 \\
\midrule
layer36 & Attention QKV & 0.25 & 0.18 & 0.19 & 0.06 & 0.16 \\
layer36 & Attention O & 0.09 & 0.06 & 0.06 & 0.03 & 0.06 \\
layer36 & FFN/MoE Gate/Up & 1.32 & 0.18 & 0.19 & 1.13 & 0.16 \\
layer36 & FFN/MoE Down & 0.64 & 0.13 & 0.14 & 0.50 & 0.11 \\
\midrule
layer40 & Attention QKV & 0.25 & 0.18 & 0.19 & 0.06 & 0.16 \\
layer40 & Attention O & 0.09 & 0.06 & 0.07 & 0.03 & 0.05 \\
layer40 & FFN/MoE Gate/Up & 1.34 & 0.19 & 0.21 & 1.14 & 0.17 \\
layer40 & FFN/MoE Down & 0.64 & 0.13 & 0.13 & 0.51 & 0.11 \\
\midrule
layer44 & Attention QKV & 0.25 & 0.19 & 0.19 & 0.06 & 0.16 \\
layer44 & Attention O & 0.09 & 0.06 & 0.07 & 0.03 & 0.06 \\
layer44 & FFN/MoE Gate/Up & 1.33 & 0.20 & 0.20 & 1.13 & 0.17 \\
layer44 & FFN/MoE Down & 0.65 & 0.13 & 0.14 & 0.51 & 0.11 \\
\bottomrule
\end{tabular}
}
\end{table}

%% file: body/checklist.tex
\section*{NeurIPS Paper Checklist}

\begin{enumerate}

\item {\bf Claims}
    \item[] Question: Do the main claims made in the abstract and introduction accurately reflect the paper's contributions and scope?
    \item[] Answer: \answerYes{} 
    \item[] Justification: All claims are supported by the experimental results.

\item {\bf Limitations}
    \item[] Question: Does the paper discuss the limitations of the work performed by the authors?
    \item[] Answer: \answerYes{} 
    \item[] Justification: Limitations are discussed in \secref{sec:conclusion}.

\item {\bf Theory assumptions and proofs}
    \item[] Question: For each theoretical result, does the paper provide the full set of assumptions and a complete (and correct) proof?
    \item[] Answer: \answerYes{} 
    \item[] Justification: Proof of theoretical results are given in \appendixref{sec:appendix-proof}.

    \item {\bf Experimental result reproducibility}
    \item[] Question: Does the paper fully disclose all the information needed to reproduce the main experimental results of the paper to the extent that it affects the main claims and/or conclusions of the paper (regardless of whether the code and data are provided or not)?
    \item[] Answer: \answerYes{} 
    \item[] Justification: The details of the implementation of our method is given in \appendixref{sec:appendix-pipeline}. All results are reproducible.

\item {\bf Open access to data and code}
    \item[] Question: Does the paper provide open access to the data and code, with sufficient instructions to faithfully reproduce the main experimental results, as described in supplemental material?
    \item[] Answer: \answerNo{} 
    \item[] Justification: The source code is currently subject to internal lab management policies regarding intellectual property review. We are committed to releasing the code and data upon paper acceptance. We have provided comprehensive pseudocode and detailed algorithm descriptions in \appendixref{sec:appendix-pipeline} to allow for independent verification and implementation.

\item {\bf Experimental setting/details}
    \item[] Question: Does the paper specify all the training and test details (e.g., data splits, hyperparameters, how they were chosen, type of optimizer) necessary to understand the results?
    \item[] Answer: \answerYes{} 
    \item[] Justification: All substantial details are provided in the paper.

\item {\bf Experiment statistical significance}
    \item[] Question: Does the paper report error bars suitably and correctly defined or other appropriate information about the statistical significance of the experiments?
    \item[] Answer: \answerNo{} 
    \item[] Justification: The runs for the experiments in the paper have low variance.

\item {\bf Experiments compute resources}
    \item[] Question: For each experiment, does the paper provide sufficient information on the computer resources (type of compute workers, memory, time of execution) needed to reproduce the experiments?
    \item[] Answer: \answerYes{} 
    \item[] Justification: All experiment setup details are listed in \secref{sec:setup}.
    
\item {\bf Code of ethics}
    \item[] Question: Does the research conducted in the paper conform, in every respect, with the NeurIPS Code of Ethics \url{https://neurips.cc/public/EthicsGuidelines}?
    \item[] Answer: \answerYes{} 
    \item[] Justification: Yes, we confirm adherence to the NeurIPS Code of Ethics.

\item {\bf Broader impacts}
    \item[] Question: Does the paper discuss both potential positive societal impacts and negative societal impacts of the work performed?
    \item[] Answer: \answerNA{} 
    \item[] Justification: Our work is focused on efficient LLM inference. We do not foresee any particular societal impacts from this work.
    
\item {\bf Safeguards}
    \item[] Question: Does the paper describe safeguards that have been put in place for responsible release of data or models that have a high risk for misuse (e.g., pre-trained language models, image generators, or scraped datasets)?
    \item[] Answer: \answerNA{} 
    \item[] Justification: This paper does not alter the capabilities of the available models or datasets, but rather provides a more efficient approach to use them on hardware with limited capabilities. Thus, we believe that our paper has a neutral risk impact in this area.

\item {\bf Licenses for existing assets}
    \item[] Question: Are the creators or original owners of assets (e.g., code, data, models), used in the paper, properly credited and are the license and terms of use explicitly mentioned and properly respected?
    \item[] Answer: \answerYes{} 
    \item[] Justification: All datasets and models used are open-source and publicly available.

\item {\bf New assets}
    \item[] Question: Are new assets introduced in the paper well documented and is the documentation provided alongside the assets?
    \item[] Answer: \answerNA{} 
    \item[] Justification: This paper does not introduce any new datasets, software libraries, or pre-trained models. All experiments are conducted using publicly available benchmarks, which are cited and described in the main paper.

\item {\bf Crowdsourcing and research with human subjects}
    \item[] Question: For crowdsourcing experiments and research with human subjects, does the paper include the full text of instructions given to participants and screenshots, if applicable, as well as details about compensation (if any)? 
    \item[] Answer: \answerNA{} 
    \item[] Justification: We did not conduct any crowdsourcing experiments or research with human subjects.

\item {\bf Institutional review board (IRB) approvals or equivalent for research with human subjects}
    \item[] Question: Does the paper describe potential risks incurred by study participants, whether such risks were disclosed to the subjects, and whether Institutional Review Board (IRB) approvals (or an equivalent approval/review based on the requirements of your country or institution) were obtained?
    \item[] Answer: \answerNA{} 
    \item[] Justification: We did not conduct any research with human subjects.

\item {\bf Declaration of LLM usage}
    \item[] Question: Does the paper describe the usage of LLMs if it is an important, original, or non-standard component of the core methods in this research? Note that if the LLM is used only for writing, editing, or formatting purposes and does \emph{not} impact the core methodology, scientific rigor, or originality of the research, declaration is not required.
    \item[] Answer: \answerNA{} 
    \item[] Justification: LLM is used only for writing, editing, and formatting purposes.

\end{enumerate}

%% file: neurips_2026.bbl
\begin{thebibliography}{10}

\bibitem{zheng2025automation}
Tianshi Zheng, Zheye Deng, Hong~Ting Tsang, Weiqi Wang, Jiaxin Bai, Zihao Wang,
  and Yangqiu Song.
\newblock From automation to autonomy: A survey on large language models in
  scientific discovery.
\newblock In {\em Proceedings of the 2025 Conference on Empirical Methods in
  Natural Language Processing}, pages 17744--17761, 2025.

\bibitem{li2024survey}
Xinyi Li, Sai Wang, Siqi Zeng, Yu~Wu, and Yi~Yang.
\newblock A survey on llm-based multi-agent systems: workflow, infrastructure,
  and challenges.
\newblock {\em Vicinagearth}, 1(1):9, 2024.

\bibitem{huang2025foundation}
Jincai Huang, Yongjun Xu, Qi~Wang, Qi~Cheems Wang, Xingxing Liang, Fei Wang,
  Zhao Zhang, Wei Wei, Boxuan Zhang, Libo Huang, et~al.
\newblock Foundation models and intelligent decision-making: Progress,
  challenges, and perspectives.
\newblock {\em The Innovation}, 6(6), 2025.

\bibitem{qwen3}
An~Yang, Anfeng Li, Baosong Yang, Beichen Zhang, Binyuan Hui, Bo~Zheng, Bowen
  Yu, Chang Gao, Chengen Huang, Chenxu Lv, et~al.
\newblock Qwen3 technical report.
\newblock {\em arXiv preprint arXiv:2505.09388}, 2025.

\bibitem{gptq}
Elias Frantar, Saleh Ashkboos, Torsten Hoefler, and Dan Alistarh.
\newblock Gptq: Accurate post-training quantization for generative pre-trained
  transformers.
\newblock {\em arXiv preprint arXiv:2210.17323}, 2022.

\bibitem{awq}
Ji~Lin, Jiaming Tang, Haotian Tang, Shang Yang, Wei-Ming Chen, Wei-Chen Wang,
  Guangxuan Xiao, Xingyu Dang, Chuang Gan, and Song Han.
\newblock Awq: Activation-aware weight quantization for on-device llm
  compression and acceleration.
\newblock {\em Proceedings of machine learning and systems}, 6:87--100, 2024.

\bibitem{efficientqat}
Mengzhao Chen, Wenqi Shao, Peng Xu, Jiahao Wang, Peng Gao, Kaipeng Zhang, and
  Ping Luo.
\newblock Efficientqat: Efficient quantization-aware training for large
  language models.
\newblock {\em arXiv preprint arXiv:2407.11062}, 2024.

\bibitem{flexgen}
Ying Sheng, Lianmin Zheng, Binhang Yuan, Zhuohan Li, Max Ryabinin, Beidi Chen,
  Percy Liang, Christopher R{\'e}, Ion Stoica, and Ce~Zhang.
\newblock Flexgen: High-throughput generative inference of large language
  models with a single gpu.
\newblock In {\em International Conference on Machine Learning}, pages
  31094--31116. PMLR, 2023.

\bibitem{hetegen}
Xuanlei Zhao, Bin Jia, Haotian Zhou, Ziming Liu, Shenggan Cheng, and Yang You.
\newblock Hetegen: Efficient heterogeneous parallel inference for large
  language models on resource-constrained devices.
\newblock {\em Proceedings of Machine Learning and Systems}, 6:162--172, 2024.

\bibitem{powerinfer}
Yixin Song, Zeyu Mi, Haotong Xie, and Haibo Chen.
\newblock Powerinfer: Fast large language model serving with a consumer-grade
  gpu.
\newblock In {\em Proceedings of the ACM SIGOPS 30th Symposium on Operating
  Systems Principles}, pages 590--606, 2024.

\bibitem{aser}
Weibo Zhao, Yubin Shi, Xinyu Lyu, Wanchen Sui, Shen Li, and Yong Li.
\newblock Aser: activation smoothing and error reconstruction for large
  language model quantization.
\newblock In {\em Proceedings of the AAAI Conference on Artificial
  Intelligence}, volume~39, pages 22822--22830, 2025.

\bibitem{eora}
Shih-Yang Liu, Maksim Khadkevich, Nai~Chit Fung, Charbel Sakr, Chao-Han~Huck
  Yang, Chien-Yi Wang, Saurav Muralidharan, Hongxu Yin, Kwang-Ting Cheng, Jan
  Kautz, et~al.
\newblock Eora: Fine-tuning-free compensation for compressed llm with
  eigenspace low-rank approximation.
\newblock {\em arXiv preprint arXiv:2410.21271}, 2024.

\bibitem{llama}
Aaron Grattafiori, Abhimanyu Dubey, Abhinav Jauhri, Abhinav Pandey, Abhishek
  Kadian, Ahmad Al-Dahle, Aiesha Letman, Akhil Mathur, Alan Schelten, Alex
  Vaughan, et~al.
\newblock The llama 3 herd of models.
\newblock {\em arXiv preprint arXiv:2407.21783}, 2024.

\bibitem{abq-llm}
Chao Zeng, Songwei Liu, Yusheng Xie, Hong Liu, Xiaojian Wang, Miao Wei, Shu
  Yang, Fangmin Chen, and Xing Mei.
\newblock Abq-llm: Arbitrary-bit quantized inference acceleration for large
  language models.
\newblock In {\em Proceedings of the AAAI Conference on Artificial
  Intelligence}, volume~39, pages 22299--22307, 2025.

\bibitem{llmint8}
Tim Dettmers, Mike Lewis, Younes Belkada, and Luke Zettlemoyer.
\newblock Gpt3. int8 (): 8-bit matrix multiplication for transformers at scale.
\newblock {\em Advances in neural information processing systems},
  35:30318--30332, 2022.

\bibitem{mobilequant}
Fuwen Tan, Royson Lee, Shell~Xu Hu, Sourav Bhattacharya, Timothy Hospedales,
  Georgios Tzimiropoulos, Brais Martinez, et~al.
\newblock Mobilequant: Mobile-friendly quantization for on-device language
  models.
\newblock {\em arXiv preprint arXiv:2408.13933}, 2024.

\bibitem{decdec}
Yeonhong Park, Jake Hyun, Hojoon Kim, and Jae~W Lee.
\newblock Decdec: A systems approach to advancing low-bit llm quantization.
\newblock In {\em 19th USENIX Symposium on Operating Systems Design and
  Implementation (OSDI 25)}, pages 803--819, 2025.

\bibitem{dlqat}
Wenjin Ke, Zhe Li, Dong Li, Lu~Tian, and Emad Barsoum.
\newblock Dl-qat: Weight-decomposed low-rank quantization-aware training for
  large language models.
\newblock {\em arXiv preprint arXiv:2504.09223}, 2025.

\bibitem{llmqat}
Zechun Liu, Barlas Oguz, Changsheng Zhao, Ernie Chang, Pierre Stock, Yashar
  Mehdad, Yangyang Shi, Raghuraman Krishnamoorthi, and Vikas Chandra.
\newblock Llm-qat: Data-free quantization aware training for large language
  models.
\newblock {\em arXiv preprint arXiv:2305.17888}, 2023.

\bibitem{accelerate}
Sylvain Gugger, Lysandre Debut, Thomas Wolf, Philipp Schmid, Zachary Mueller,
  Sourab Mangrulkar, Marc Sun, and Benjamin Bossan.
\newblock Accelerate: Training and inference at scale made simple, efficient
  and adaptable.
\newblock \url{https://github.com/huggingface/accelerate}, 2022.

\bibitem{deepspeed}
Reza~Yazdani Aminabadi, Samyam Rajbhandari, Ammar~Ahmad Awan, Cheng Li, Du~Li,
  et~al.
\newblock Deepspeed-inference: enabling efficient inference of transformer
  models at unprecedented scale.
\newblock In {\em SC22: International Conference for High Performance
  Computing, Networking, Storage and Analysis}, pages 1--15. IEEE, 2022.

\bibitem{fiddler}
Keisuke Kamahori, Tian Tang, Yile Gu, Kan Zhu, and Baris Kasikci.
\newblock Fiddler: Cpu-gpu orchestration for fast inference of
  mixture-of-experts models.
\newblock {\em arXiv preprint arXiv:2402.07033}, 2024.

\bibitem{hobbit}
Peng Tang, Jiacheng Liu, Xiaofeng Hou, Yifei Pu, Jing Wang, Pheng-Ann Heng,
  Chao Li, and Minyi Guo.
\newblock Hobbit: A mixed precision expert offloading system for fast moe
  inference.
\newblock {\em arXiv preprint arXiv:2411.01433}, 2024.

\bibitem{lqlora}
Han Guo, Philip Greengard, Eric~P Xing, and Yoon Kim.
\newblock Lq-lora: Low-rank plus quantized matrix decomposition for efficient
  language model finetuning.
\newblock {\em arXiv preprint arXiv:2311.12023}, 2023.

\bibitem{loftq}
Yixiao Li, Yifan Yu, Chen Liang, Pengcheng He, Nikos Karampatziakis, Weizhu
  Chen, and Tuo Zhao.
\newblock Loftq: Lora-fine-tuning-aware quantization for large language models.
\newblock {\em arXiv preprint arXiv:2310.08659}, 2023.

\bibitem{lorc}
Rongzhi Zhang, Kuang Wang, Liyuan Liu, Shuohang Wang, Hao Cheng, Chao Zhang,
  and Yelong Shen.
\newblock Lorc: Low-rank compression for llms kv cache with a progressive
  compression strategy.
\newblock {\em arXiv preprint arXiv:2410.03111}, 2024.

\bibitem{huggingface}
Thomas Wolf, Lysandre Debut, Victor Sanh, Julien Chaumond, Clement Delangue,
  Anthony Moi, Pierric Cistac, Tim Rault, R{\'e}mi Louf, Morgan Funtowicz,
  et~al.
\newblock Transformers: State-of-the-art natural language processing.
\newblock In {\em Proceedings of the 2020 conference on empirical methods in
  natural language processing: system demonstrations}, pages 38--45, 2020.

\bibitem{llamacpp}
G.~Gerganov and contributors.
\newblock llama.cpp: Inference of llama model in pure c/c++.
\newblock \url{https://github.com/ggml-org/llama.cpp}, 2025.

\bibitem{vllm}
Woosuk Kwon, Zhuohan Li, Siyuan Zhuang, Ying Sheng, Lianmin Zheng, Cody~Hao Yu,
  Joseph Gonzalez, Hao Zhang, and Ion Stoica.
\newblock Efficient memory management for large language model serving with
  pagedattention.
\newblock In {\em Proceedings of the 29th symposium on operating systems
  principles}, pages 611--626, 2023.

\bibitem{arc}
Peter Clark, Isaac Cowhey, Oren Etzioni, Tushar Khot, Ashish Sabharwal, Carissa
  Schoenick, and Oyvind Tafjord.
\newblock Think you have solved question answering? try arc, the ai2 reasoning
  challenge.
\newblock {\em arXiv preprint arXiv:1803.05457}, 2018.

\bibitem{mathqa}
Aida Amini, Saadia Gabriel, Shanchuan Lin, Rik Koncel-Kedziorski, Yejin Choi,
  and Hannaneh Hajishirzi.
\newblock Mathqa: Towards interpretable math word problem solving with
  operation-based formalisms.
\newblock In {\em Proceedings of the 2019 Conference of the North American
  Chapter of the Association for Computational Linguistics: Human Language
  Technologies, Volume 1 (Long and Short Papers)}, pages 2357--2367, 2019.

\bibitem{mmlu}
Dan Hendrycks, Collin Burns, Steven Basart, Andy Zou, Mantas Mazeika, Dawn
  Song, and Jacob Steinhardt.
\newblock Measuring massive multitask language understanding.
\newblock {\em arXiv preprint arXiv:2009.03300}, 2020.

\bibitem{c4}
Jesse Dodge, Maarten Sap, Ana Marasovi{\'c}, William Agnew, Gabriel Ilharco,
  Dirk Groeneveld, Margaret Mitchell, and Matt Gardner.
\newblock Documenting large webtext corpora: A case study on the colossal clean
  crawled corpus.
\newblock In {\em Proceedings of the 2021 conference on empirical methods in
  natural language processing}, pages 1286--1305, 2021.

\end{thebibliography}
